%% file: 0_main.tex
\documentclass[fleqn,10pt]{wlscirep}
\usepackage[utf8]{inputenc}
\usepackage[T1]{fontenc}
\newcommand\blfootnote[1]{%
  \begingroup
  \renewcommand\thefootnote{}\footnote{#1}%
  \addtocounter{footnote}{-1}%
  \endgroup
}
\title{Efficient Adaptation of Deep Neural Networks for Semantic Segmentation in Space Applications}

\author[1]{Leonardo Olivi}
\author[2]{Edoardo Santero Mormile}
\author[3]{Enzo Tartaglione}

\affil[1]{Freelance,\tt leonardo.olivi.98@gmail.com}
\affil[2]{University of Trento, Italy, \tt e.santeromormile@unitn.it}
\affil[3]{LTCI, Télécom Paris, Institut Polytechnique de Paris, France,\tt enzo.tartaglione@telecom-paris.fr}

\newcommand{\eg}{\emph{e.g.}}
\newcommand{\etal}{\emph{et~al.~}}
\newcommand{\R}{\mathbb{R}}
\newcommand{\BN}{\text{BN}}

\newcommand{\beginsupplement}{%
        \setcounter{table}{0}
        \renewcommand{\thetable}{ST\arabic{table}}%
        \setcounter{figure}{0}
        \renewcommand{\thefigure}{SF\arabic{figure}}%
        \setcounter{section}{1}
        \renewcommand{\thesection}{S\arabic{section}}%
        }

\usepackage{bbold}
\usepackage{adjustbox}
\usepackage{multirow}
\usepackage{subcaption}
\usepackage{wrapfig}

\begin{abstract}
In recent years, the application of Deep Learning techniques has shown remarkable success in various computer vision tasks, paving the way for their deployment in extraterrestrial exploration. Transfer learning has emerged as a powerful strategy for addressing the scarcity of labeled data in these novel environments. This paper represents one of the first efforts in evaluating the feasibility of employing adapters toward efficient transfer learning for rock segmentation in extraterrestrial landscapes, mainly focusing on lunar and martian terrains. Our work suggests that the use of adapters, strategically integrated into a pre-trained backbone model, can be successful in reducing both bandwidth and memory requirements for the target extraterrestrial device. In this study, we considered two memory-saving strategies: layer fusion (to reduce to zero the inference overhead) and an ``adapter ranking'' (to also reduce the transmission cost). Finally, we evaluate these results in terms of task performance, memory, and computation on embedded devices, evidencing trade-offs that open the road to more research in the field. 
\end{abstract}

\begin{document}

\flushbottom
\maketitle
\thispagestyle{empty}

\input{1_introduction}


\input{2_related_work}


\input{3_adapters}


\input{4_experiments}


\input{5_discussion}


\bibliography{bibliography}


\section*{Acknowledgements}

This paper has been supported by the French National Research Agency (ANR) in the framework of
the JCJC project ``BANERA'' ANR-24-CE23-4369, and by Hi!PARIS Center on Data Analytics and Artificial Intelligence. 


\section*{Author contributions statement}

E.T. conceived the experiment(s),  L.O. and E.S.M. conducted the experiment(s), L.O. and E.S.M. analyzed the results. All authors reviewed the manuscript.


\section*{Additional information}
\textbf{Competing interests.} The corresponding author is responsible for submitting a \href{http://www.nature.com/srep/policies/index.html#competing}{competing interests statement} on behalf of all authors of the paper.


\newpage
\beginsupplement
\input{supplementary}


\end{document}

%% file: 1_introduction.tex
\section{Introduction}
\begin{wrapfigure}{r}{0.5\textwidth}
  \begin{center}
    \includegraphics[width=0.48\textwidth]{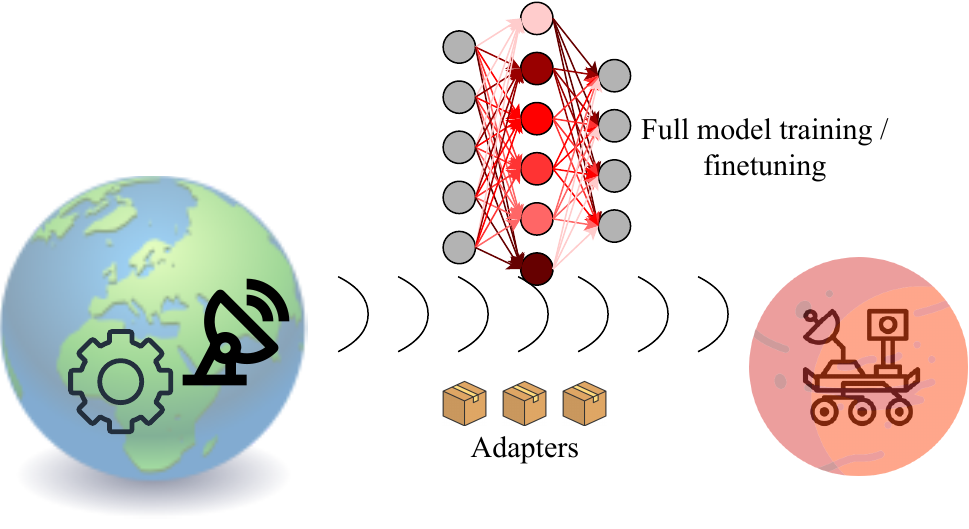}
  \end{center}
  \caption{Transmitting a fully fine-tuned model can be very bandwidth and memory-demanding. In this work, we explore the possibility of employing adapters, shallow modules that introduce corrections to a pre-loaded one (and that can be considered as an update).}
  \label{fig:teaser}
\end{wrapfigure}
The\blfootnote{This work has been accepted for publication at Scientific Reports.} exploration of celestial bodies beyond Earth requires deploying sophisticated computer vision systems endowed with the capability to robustly identify and characterize surface features~\cite{ge2019recent}. Among the critical tasks in this extraterrestrial context, rock segmentation plays a critical role, since the geological composition of these distant terrains offers vital insight into the history and potential habitability of celestial bodies~\cite{gor2001autonomous,furlan2019rock}. However, the acquisition of labeled data for training machine learning models in such environments proves to be an arduous challenge, primarily due to the impracticality of physical presence and the associated complexities of data collection. 

Another obstacle lies in the difficulty of transmitting large volumes of data from remote celestial bodies back to Earth~\cite{gor2001autonomous}. The limited bandwidth and communication capabilities of space missions of impose constraints on the efficiency and feasibility of transmitting extensive datasets. Consequently, this limitation hinders the traditional approach of gathering abundant labeled data for training and updating machine learning models deployed in extraterrestrial environments. 

Besides, the dynamic nature of extraterrestrial landscapes, subject to geological and environmental variations, underscores the need for adaptive and updatable models. Continuous updates to machine learning models become imperative to account for changes in surface features, ensuring that the algorithms remain effective and accurate over extended mission durations. Efficient transmission of these updates becomes a critical consideration as it involves optimizing the use of limited bandwidth to relay pertinent model adjustments without compromising the overall mission (see Fig.~\ref{fig:teaser}).

In response to the formidable challenge of acquiring labeled data for training machine learning models in extraterrestrial environments, we present a pioneering exploratory study leveraging the concept of \textit{adapters} applied to a pre-trained backbone~\cite{rebuffi2017learning} for space applications and efficient memory-saving strategies. Concretely, we contribute in three different aspects. 
\begin{itemize}[noitemsep]
    \item Definition of the structure of adapters properly suited for this task (Sec.~\ref{sec:adapters}). Although many designs have been proposed in recent years, we customize the design to be adaptable to the realistically deployable architectures, reducing transmission and inference costs.
    \item Analyze the real memory complexity for the inserted modules (Sec.~\ref{sec:memcomplex}). This aspect covers a marginal role in other works where adapters are proposed, but in our design, the memory footprint covers a central role in the solution's deployability. 
    \item Implementation of two strategies that enable the deployability of the proposed method: \emph{adapter fusion}, where we explain how to fuse the adapters to the original backbone, causing no computational overhead (Sec.~\ref{sec:adapter_fusion}), and \emph{adapter ranking}, allowing us to selectively choose which layers require adaptation, saving bandwidth for the update (Sec.~\ref{sec:adapt_ranks}).
\end{itemize}
We validate our study through a quantitative evaluation of the task of interest, evaluating trade-offs in terms of the size of the VS model in performance and the computation required, showing that adapters provide a fair compromise between performance and memory size (Sec.~\ref{sec:experiments}). This study is also completed by measures on real embedded devices, either GPU equipped or not.

%% file: 2_related_work.tex
\section{Related Work}
\label{sec:related_work}

\subsection{Traditional Approaches to Rock Segmentation}
The task of rock segmentation in extraterrestrial environments deepens its roots far back to the past. The work by Gor~\etal provided one of the first possible setups that allowed autonomous rock segmentation for Mars rovers~\cite{gor2001autonomous}. Specifically, in this work, the authors introduce a parameter-independent framework for autonomous rock detection in Martian terrain, with a demonstrated algorithm applied to real Mars Rover data. The idea behind this work was to build foundations for all the upcoming work: facilitate efficient decision making, aid in prioritizing data for transmission, select regions for in-depth scientific measurements, and optimize exploration strategies to maximize scientific returns per transmitted data bit. This was achieved through clustering algorithms. Thompson and Castrono performed a first comparison between seven segmentation algorithms (including Support Vector Machines) was performed by Thompson~and~Casta\~no~\cite{thompson2007performance}, where the best overall performance was attributed to the \emph{Rockfinder} algorithm~\cite{castano2005current} having target selection accuracy rates~\cite{castano2005current} of approximately the~80\%, but highlighting as well that finding all rocks in the image is a difficult task, obtaining a 60\% recall. Other approaches in later years tried to improve performance on this task, including the use of random forests to perform semantic segmentation for hazard detection on planetary rovers~\cite{ono2015risk}, edge regrouping~\cite{burl2016rockster}, the exploitation of region contrast to aid in segmentation~\cite{xiao2017autonomous}, superpixel graph cuts~\cite{gong2012rock} and including size-frequency priors in the fitting analysis~\cite{golombek2012detection}. 

\subsection{Deep Learning for Rock Segmentation}
Although traditional approaches are a better fit for working in aerospace environments where computation and memory are very limited, the performance of Deep Learning (DL) approaches shows that moving towards these new techniques is currently the most promising direction despite the high computational costs~\cite{furlan2019rock}. The first pioneering approach tried to overcome low generalization performance with the employment of Support Vector Machines~\cite{dunlop2007multi}: since traditional visual segmentation techniques struggle with the diverse morphologies of rocks, this work integrates features from multiple scales. More recently, several recent advancements in rock segmentation for extraterrestrial environments have been proposed, showcasing diverse approaches to address the challenges of navigation and classification in extraterrestrial environments. Kuang~\emph{et~al.}~\cite{kuang2021rock} employ a variant of U-Net++~\cite{zhou2019unet} to achieve rock segmentation for Mars navigation systems. Deep Mars~\cite{WagstaffLSGGP18} trains an AlexNet to classify engineering-focused rover images, albeit limited to recognizing a single object per image. The Soil Property and Object Classification~\cite{rothrock2016spoc} introduces a fully convolutional neural network for Mars terrain segmentation. Swan~\emph{et~al.}~\cite{AI4Mars} contribute a terrain segmentation dataset and assess performance using DeepLabv3+~\cite{DeeplabV3_plus}. Notably, Transformer-based networks have emerged as a focus, with studies such as RockFormer \cite{liu2023rockformer} and MarsFormer~\cite{xiong2023marsformer} demonstrating their efficacy in Martian rock segmentation. Addressing broader visual navigation challenges, Zhang~\etal\cite{zhang2018novel} tackles Mars' visual navigation problem through a deep neural network capable of finding optimal paths in the global Martian environment. Finally, the space hardware limitations, spanning across ensuring their memory and computational efficiency, constrain heavily the DL implementations, requiring a careful and deepened study of the optimization of these, Marek~\etal\cite{MAREK2024107311}.

\subsection{Transfer Learning Approaches for Deep Learning}
In the context of future extraterrestrial environments, the pivotal role of DL is anticipated; however, its effectiveness is impeded by the inherent challenge of acquiring annotated training data. Traditional machine learning algorithms often presuppose that training and testing data must inhabit the same feature space and exhibit similar distributions~\cite{pan2009survey}, a condition that proves challenging in the context of obtaining an adequate quantity of Martian rock images. Addressing this, knowledge transfer emerges as a promising avenue to enhance learning performance and alleviate the challenges associated with dataset acquisition in extraterrestrial scenarios. Li~\etal\cite{li2020autonomous} categorize transfer learning approaches into four distinct scenarios. Firstly, \textit{parameter-transfer} involves identifying shared parameters or prior distributions between the source and target domain models, aiming to enhance learning performance~\cite{sargano2017human}. Secondly, \textit{feature-representation-transfer} focuses on learning representative features through the source domain, with the encoded knowledge transferred across domains embedded within the learned feature representation~\cite{zhao2018transfer,quetu45dsd2}. The \textit{relational-knowledge-transfer} scenario assumes relational and independently identically distributed source and target domains, aiming to establish a mapping of relational knowledge between the two~\cite{wang2016relational}. Lastly, \textit{instance-based transfer} involves the reuse of a portion of data from the source domain for learning in the target domain through a process of reweighting~\cite{wang2014new,wang2019instance}. These categorizations provide a comprehensive overview of the diverse strategies employed in transfer learning scenarios. In our work we will be falling into the parameter-transfer scenario, adapting the parameter's distribution efficiently with the employment of a solution inspired by adapters~\cite{rebuffi2017learning}. Such an approach also shows further potential to reduce the model's complexity, employing pruning at the adapter's scale~\cite{marouf2024mini}.

\subsection{Challenges of Deploying Neural Networks in Space Environments}
Despite such a high level of performance, deploying neural network-based models for image segmentation in space exploration presents unique challenges due to the constrained operational environment~\cite{furano2020towards}. Space-borne systems operate with significantly limited computational resources compared to terrestrial applications, as space-qualified hardware has lower processing power and memory~\cite{ziaja2021benchmarking,nguyenactivation}, necessitating model optimization techniques such as quantization, pruning, and adapter-based fine-tuning \cite{hanetal2016,Najafabadi2015DeepLA,dhar2021survey,rajapakse2023intelligence}. Communication constraints further complicate deployment, as deep space missions suffer from high-latency and low-bandwidth transmission, making real-time updates impractical. Autonomous models must be capable of in-situ learning and quick adaptation to maintain performance under changing environmental conditions or rapidly evolving environments~\cite{chien_castillo2014}. Furthermore, due to such a quick and active environment, these models need to be robust to noise and malfunctions, necessitating fault-tolerant AI implementations. Understanding and quantifying the impact of such issues on DL models deployed remains under-explored\cite{Nalepaetal2021}, but it is a challenge to address in this field. Solutions like providing a custom design relying on annotated images~\cite{bamford2021deep}. These constraints underscore the necessity of designing neural networks that are efficient and adaptable, particularly for image segmentation tasks in space exploration scenarios.

%% file: 3_adapters.tex
\section{Efficient Adaptation of a DL model}
\label{sec:method}

\begin{wrapfigure}{r}{0.5\textwidth}
    \vspace{-10pt}
    \centering
    \includegraphics[scale=0.38]{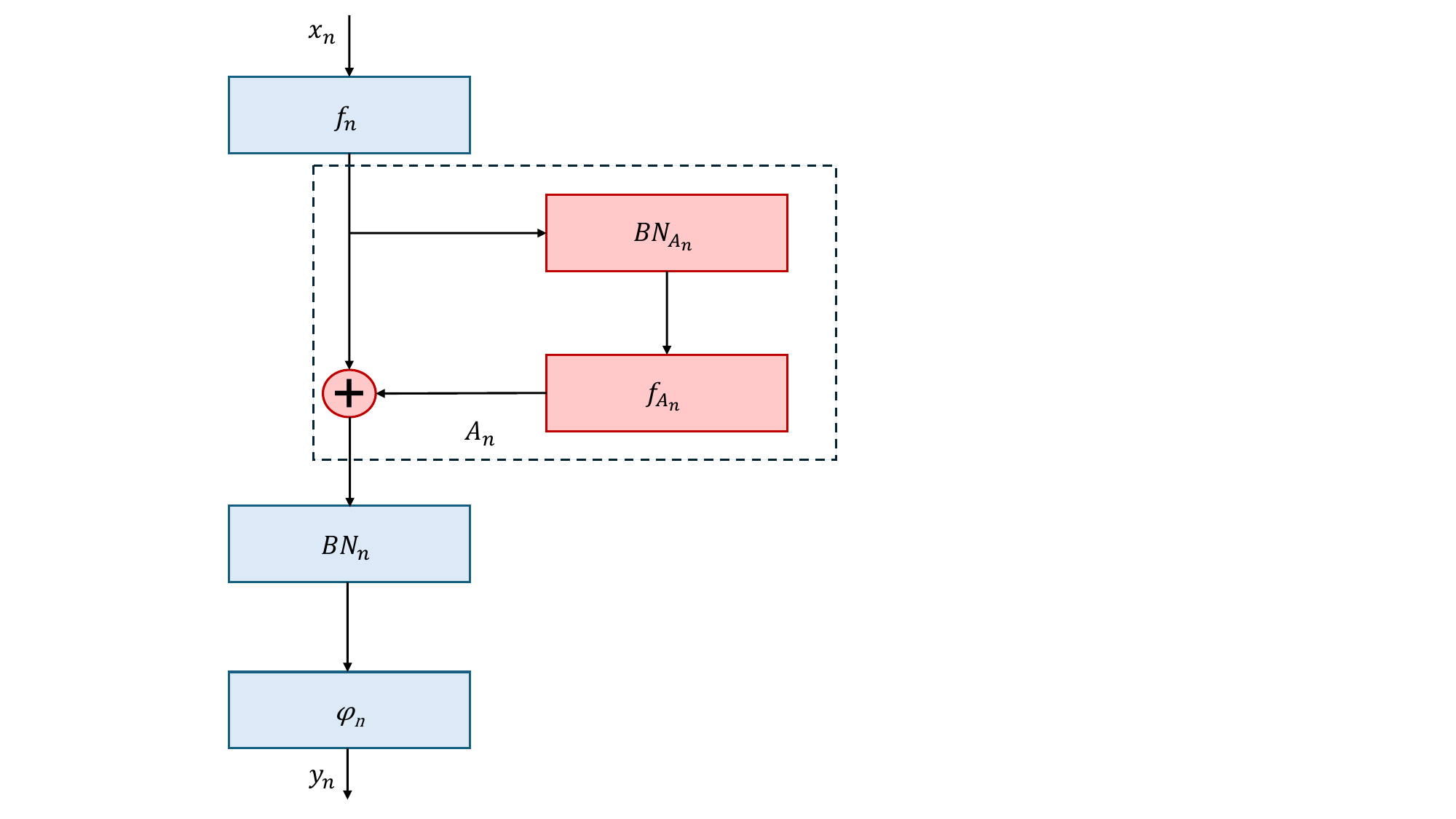}
    \caption{A schematic showing how adapters ($A_n$, in red) are plugged into the rest of the architecture (blue).}
    \label{fig:adapter}
\end{wrapfigure}

In this section, we will provide an overview of the techniques chosen to make a Deep Neural Network (DNN) able to work efficiently in different environments, namely by using adapters. We motivate their choice for our specific setup, their architecture, and implementation, together with our \emph{adapter fusion} and \emph{adapter ranking}.

\subsection{Adapters Design}
\label{sec:adapters}
The main inspiration for the adapters' injection comes from the work by Rebuffi~\etal work\cite{adaptersold} concerning residual adapters originally introduced in a ResNet-15 model. Within the cited works, the design of residual adapters has shown great performance, outperforming standard transfer learning techniques. Let $y_n$ be the output of the $n$-th layer in a DNN model. We can generally model layers in recent popular models like ResNets and even more traditional ones like VGGs can be expressed as
\begin{equation}
    y_n = \varphi_n\left\{{\BN}_n\left[f_n\left(x_n, w_{f_n}\right), w_{\BN_n}\right]\right\},
\end{equation}
where $\varphi_n$ is the applied non-linearity, $\BN_n$ is the Batch normalization layer, $f_n$ is the linear transformation applied (\eg ~fully connected or convolutional), and $w_{f_n}$ and $ w_{\BN_n}$ are the learnable parameters associated to $f_n$ and $\BN_n$, respectively.

Let us assume our model is trained on some upstream task $T_0$ and should be fine-tuned on a downstream task $T_1$: in the case we update the parameters of the $n$-th layer, we have
\begin{equation}
	\label{eq:regularforw}
	y_n^{T_1} = \varphi_n\left\{\BN_n\left[f_n\left(x_n,  w_{x}^{T_1}\right), w_{\BN_n}^{T_1}\right]\right\}
\end{equation}
having
\begin{equation}
	w_x^{T_1} = w_{x}^{T_0}+\Delta w_{x}^{T_1},
\end{equation}
where $ w_{x}^{T_0}$ indicates the original parameters from $T_0$ and $\Delta w_{x}^{T_1}$ is the parameter's change after fine-tuning on $T_1$. If $T_1$ is very correlated with $T_0$, we can expect that $\Delta w_{x}^{T_1}\approx 0$, meaning that its dimensionality can be reduced. Hence, instead of learning $w_{x}^{T_1}$ initializing it to $w_{x}^{T_0}$, we can learn $\Delta w_{x}^{T_1}$ directly: we can inject then, between $f_n$ and $\BN_n$, a \emph{series adapter}
\begin{equation}
	A_n^{T_1} = f_{A_n}\left[ f_{n}\left(x_n, w_{f_n}^{T_0}\right), w_{f_{A_n}}^{T_1}\right]
\end{equation}
entitled to learn the parameters' shift for $T_1$ implicitly and to be directly plugged into the DNN. At this point, \eqref{eq:regularforw} writes
\begin{equation}
    y_n^{T_1} = \varphi_n \left\{{\BN}_n\left[A_n^{T_1}\left(f_n\left(x_n,  w_{x}^{T_0}\right), w_{f_{A_n}}^{T_1}\right) + f_n\left(x_n, w_{x}^{T_0}\right), w_{\BN_n}^{T_0}\right] \right\}.
 \label{eq:Tforw}
\end{equation}
In our design, we want $f_{A_n}$ to maintain the same output dimensionality; hence good choices for it are $1\times 1$ convolutions or Batch Normalizations; we choose for our design to employ both:
\begin{equation}
	\label{eq:adapter}
	A_n = f_{A_n}\left\{\BN_{A_n}\left[f_n\left(x_n, w_{f_n}^{T_0}\right), w_{\BN_{A_n}}\right], w_{f_{A_n}}\right\}.
\end{equation}
A visual representation of \eqref{eq:adapter} can be found in Fig.~\ref{fig:adapter}.

\subsection{Memory Complexity of Adapters}
\label{sec:memcomplex}
Let us analyze the complexity of employing the Adapters structure described in Sec.~\ref{sec:adapters}. Let us take the typical case where adapters are employed to correct convolutional layers. Let us say $w_{f_n}\in \R^{K_n \times K_n \times I_n \times O_n}$, where $K_n \times K_n \times I_n$ is the single filter size and $O_n$ is the number of filters in the $n$-th layer. Please note that an extra $O_n$ term may come from biases that are typically absent if batch normalization, as in our case, follows the convolutional layer. 

We know that the space complexity of this layer is $\Theta(K_n \times K_n \times I_n \times O_n)$. In our design, the adapters are constituted of a batch normalization layer having $\Theta(4 \cdot O_n)$ (as it includes a set of four parameters per channel) and the $1\times 1$ convolution layer has $\Theta(O_n)$. Overall, we know that each adapter has space complexity $\Theta(5\cdot O_n)$ of parameters to be updated, in place of  $\Theta[O_n(K_n \times K_n \times I_n+5)]$ (that include both $f_n$ and $\BN_n$): evidently, for $K_n>1$ and $I_n>1$ (which is the typical case) we have a considerable parameter complexity saving, even at train time. This comes, however, at the cost of extra computation at inference time. In principle, this is avoidable by layer folding, possible given that in the proposed design we have included no non-linearity between $f_n$ and $A_n$: this is left for further studies.

\subsection{Adapter fusion} 
\label{sec:adapter_fusion}
\begin{figure}[ht]
\begin{subfigure}[b]{0.25\textwidth}
     \centering
     \includegraphics[scale=0.3]{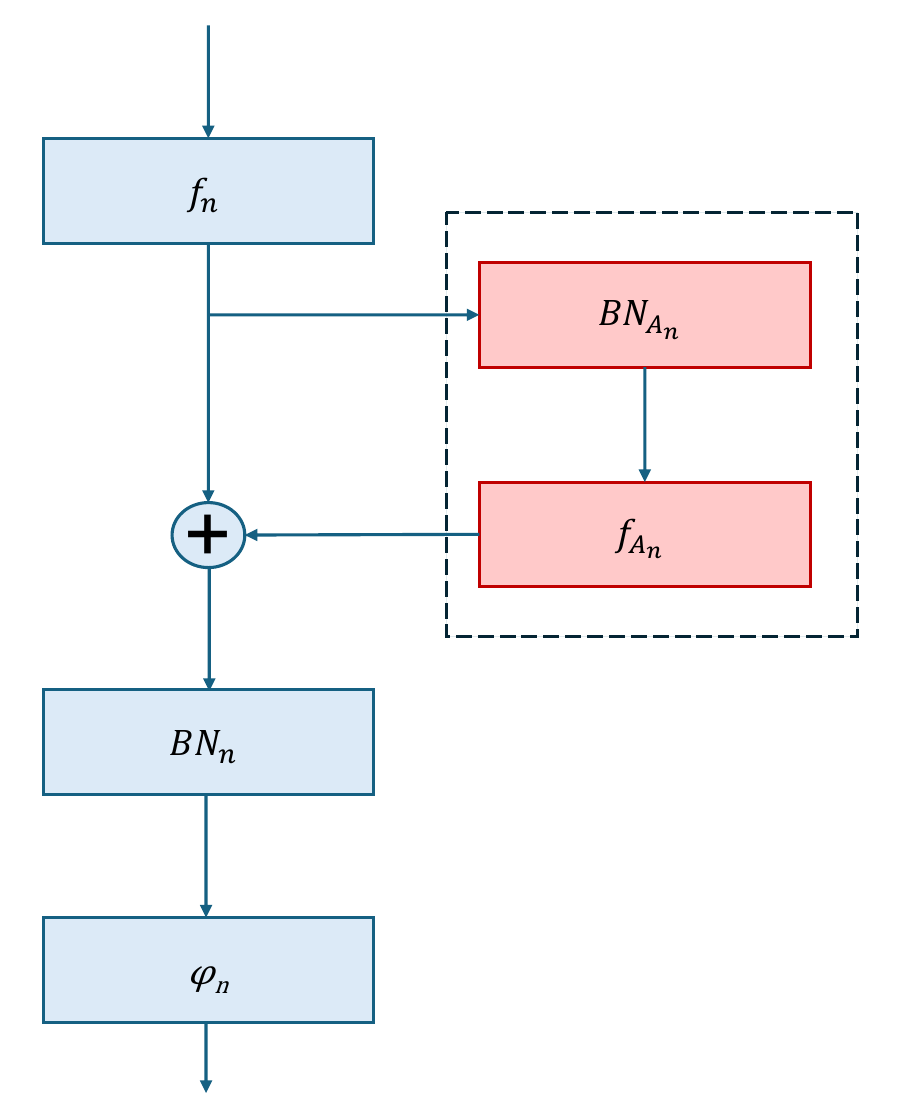}
     \caption{Step 1.}
\end{subfigure}
\hspace{0.2cm}
\begin{subfigure}[b]{0.25\textwidth}
     \centering
     \includegraphics[scale=0.3]{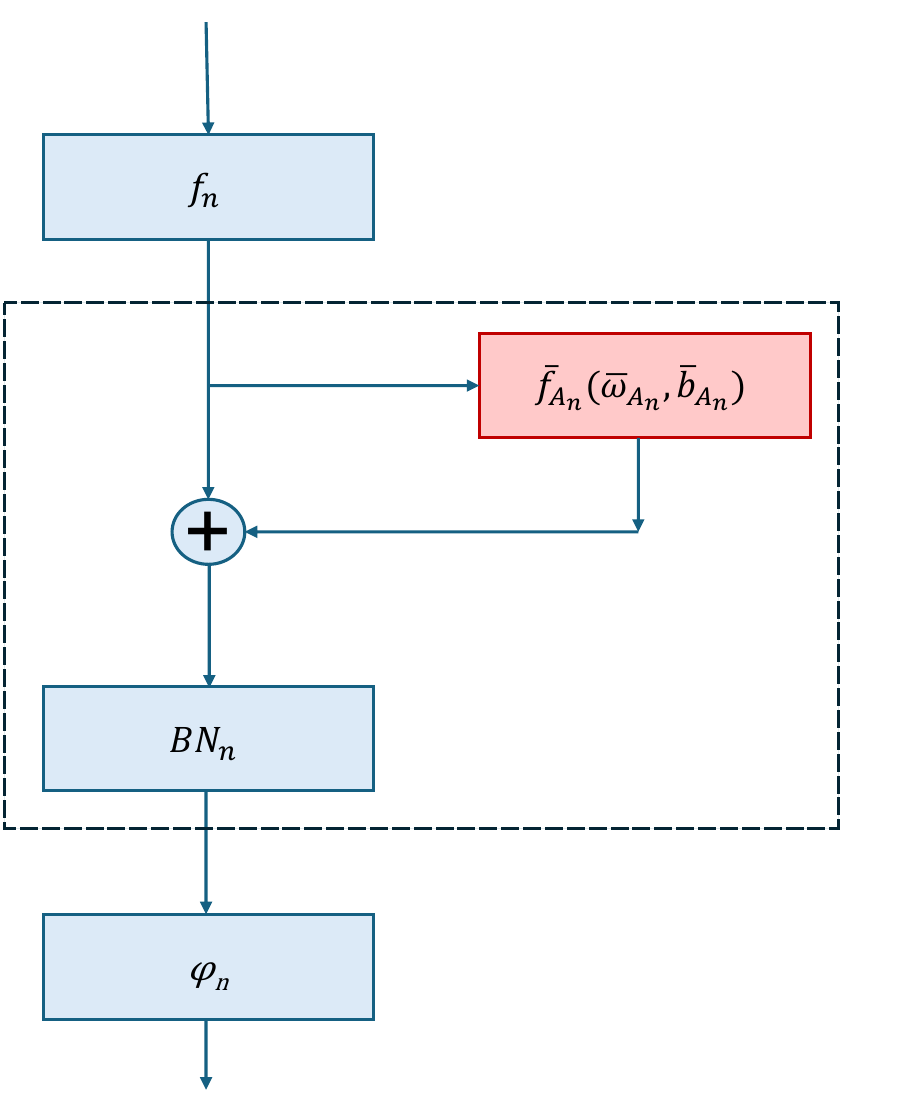}
     \caption{Step 2.}
\end{subfigure}
\hspace{-0.2cm}
\begin{subfigure}[b]{0.25\textwidth}
     \centering
     \raisebox{0.5cm}{
     \includegraphics[scale=0.3]{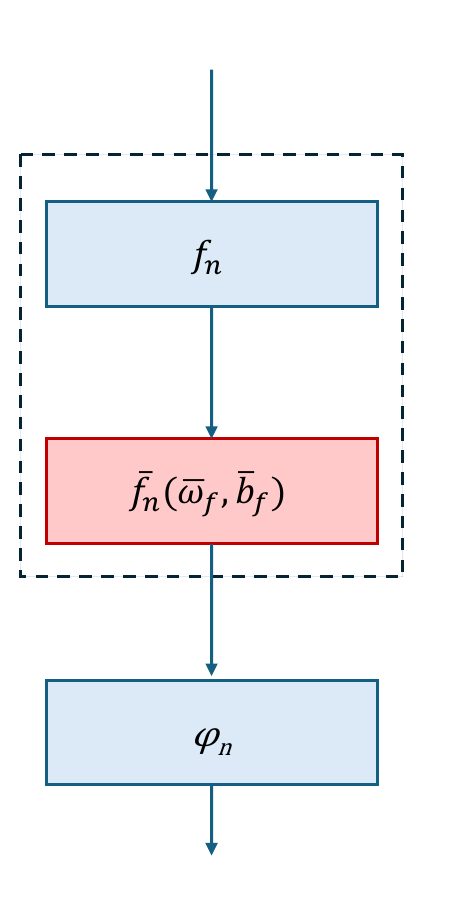}}
     \caption{Step 3.}
\end{subfigure}
\hspace{-0.5cm}
\begin{subfigure}{0.25\textwidth}
    \centering
    \raisebox{1cm}{
    \includegraphics[scale=0.3]{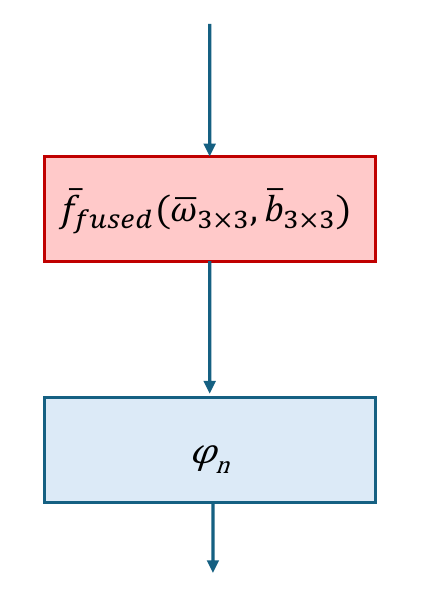}}
    \caption{Final result.}
\end{subfigure}
\caption{Adapter fusion scheme, visualized step-by-step.}
\label{fig:fusemethod}
\end{figure}
A way to decrease memory consumption and, simultaneously, the computational cost is to compress the architecture by fusing the adapter batch normalization layers and the sum operation with the adapter $1\times 1$ convolutional layer (Step 1). After the adapter is fused, the newly obtained $1\times 1$ convolutional layer is fused with the batch norm layer of the original architecture (Step 2). Finally, the $1\times 1$ convolutional layer is fused into the backbone's original $3\times 3$ convolutional layer (Step 3).  Fig.~\ref{fig:fusemethod} visualizes how to fuse these layers step-by-step. 

\textbf{Step 1.}
For the sake of clarity, we simplify the adapter operation defined in \eqref{eq:adapter} to focus on the important aspects of the compression. Let $f_n$ be the output of the n-th layer before the adapter, $f_{A_n}$ be a $1\times 1$ convolutional operation, and $\BN_{A_n}$ be a batch norm operation. The output of the layer following the adapter, indicated as $\tilde{f_n}$, is equal to
\begin{equation}
	\tilde{f_n} = f_n \ + \ f_{A_n}\left[\BN_{A_n}(f_n)\right].
    \label{eq:overallada}
\end{equation}
We will start by detailing the matrix operations behind the batch norm and convolutional layers. Let $w_{\BN_{A_n}}$ and $b_{\BN_{A_n}}$ be weights and biases of shape $ \R^{C}$, where $C$ is the number of features of previous layers. The batch norm layer computes the running mean $\mathbb{E}[x]$, and running standard deviation $\sigma[x]$, where $x$ is the output of the previous layer. Then, it performs a scaling and an element-wise multiplication, which we indicate with the symbol $\odot$. The forward operation of a batch norm layer can be defined as
\begin{equation}
    \BN(x) = \frac{x - \mathbb{E}[x]}{\sigma[x]} \odot w_{\BN_{A_n}} + b_{\BN_{A_n}}.
    \label{eq:BN}
\end{equation}
For the convolutional layers, we use $*$ to indicate the convolutional operation, $w_{f_{A_n}}$ convolutional weights and $b_{f_{A_n}}$ the bias. With $x$ being the output of previous layers and $f$ indicating the convolutional operation, we can define the convolutional forward operation as
\begin{equation}
    f(x) = w_{f_{A_n}} * x + b_{f_{A_n}}.
    \label{eq:conv}
\end{equation}
We can now put all the operations together: replacing \eqref{eq:BN} and \eqref{eq:conv} in \eqref{eq:overallada}, we have
\begin{equation}
	\tilde{f_n} = f_n + w_{f_{A_n}} * \left[\frac{f_n -\mathbb{E}[f_n]}{\sigma[f_n]} \odot w_{\BN_{A_n}} + b_{\BN_{A_n}}\right] + b_{f_{A_n}}.
\end{equation}
Now, let $\mathbb{1}$ be an identity matrix of shape equal to $w_{A_n}$ and $\times$ the matrix multiplication symbol. We can now define two equivalent weights and biases to compress our operation, which will be indicated respectively as $\tilde{w}_{A_n}$ and $\tilde{b}_{A_n}$
\begin{align}
	\tilde{w}_{A_n} &= \mathbb{1} + w_{f_{A_n}} \odot \frac{w_{\BN_{A_n}}}{\sigma[f_n]}\nonumber\\
	\tilde{b}_{A_n} &= w_{\BN_{A_n}} \times \left(b_{\BN_{A_n}} - \frac{w_{\BN_{A_n}} \odot\mathbb{E}[f_n]}{\sigma[f_n]} \right)+ b_{f_{A_n}}.
\end{align}

\textbf{Step 2.}
Now, after the adapter compression, we find ourselves with a $3 \times 3$ convolutional layer, a $1 \times 1$ convolutional layer, and a batch norm layer. To avoid confusion, let us indicate $x$ as the output of the $3\times 3$ convolutional operation and $f$ as the $1 \times 1$ convolutional operation. We use $w_f$ and $b_f$ for the convolutional weights and biases, $w_\text{BN}$ and $b_\text{BN}$ for the batch norm weights and biases, $\mathbb{E}[f(x)]$ and $\sigma[f(x)]$ for running averages and standard deviations. The sequence of the two operations yields
\begin{equation}
    \text{BN}(f(x)) = \frac{w_f \odot w_\text{BN}}{\sigma[f(x)]} \odot x + w_\text{BN} \times \frac{b_f -\mathbb{E}[f(x)]}{\sigma[f(x)]} + b_\text{BN}.
\end{equation}
Therefore, we can define new fused weights and biases as
\begin{equation}
     \tilde{w}_f = \frac{w_f \odot w_\text{BN}}{\sigma[f(x)]},~~~~~~~~~~~~~~~~~~~~~~~~~~~~~~~~~~~~~~~~~~~~~~~~~~~~~~~~~~~~~~~~~~~~~~~~~~~~\tilde{b}_f = w_\text{BN} \times \frac{b_f -\mathbb{E}[f(x)]}{\sigma[f(x)]} + b_\text{BN}.
\end{equation}

\textbf{Step 3.}
Finally, we have a $3 \times 3$ convolutional layer and a $1 \times 1$ convolutional layer. Weights and biases of the former are represented by $w_{3\times 3}$ and $b_{3\times 3}$, whereas the latter ones are $w_{1\times 1}$ and $b_{1\times 1}$. We indicate with $C_\text{out}$ the output channel and $C_\text{in}$ the input channel. Therefore, we have $w_{3\times 3} \in \R^{C_\text{out} \times C_\text{in} \times 3 \times 3}$ and $b_{3\times 3}\in \R^{C_\text{out}}$, while $w_{1\times 1}\in \R^{C_\text{out} \times C_\text{in}}$ and $b_{1\times 1}\in \R^{C_\text{out}}$.
Each output channel $o \in [0,C_{out}-1]$ of the fused weight $\tilde{w}_{3\times 3}$ and bias $\tilde{b}_{3\times 3}$ can be computed as
\begin{equation}
    \tilde{w}_{3\times 3}[o,:,:,:] = \sum_{m=0}^{C_{out}} w_{1x1}[o,m] \cdot w_{3x3}[m,:,:,:],~~~~~~~~~~~~~~~~~~~~~~~~~
    \tilde{b}_{3\times 3}[o] = b_{1x1}[o] + \sum_{m=0}^{C_{out}} w_{1x1}[o,m] \cdot b_{3x3}[m].
\end{equation}

This ``fusion'' procedure helps the adapted-equipped architecture, eliminating the operational cost of the adapter introduction, namely the new layers insertion. It happens in two ways: first of all, the adapter fusion reduces the total number of layers, lowering the number of operations (i.e. FLOPs) of the model and eliminating the FLOPs related to the adapter insertion. Furthermore, the adapter's size is absorbed within the architecture, zeroing its memory size cost. Finally, performing the fusion returns the model structure to the starting one, making new modifications feasible and simpler.

\subsection{Adapters Ranking} 
\label{sec:adapt_ranks}
In this section, we propose a method for ranking adapters and assessing whether they are worth keeping in the architecture or can be discarded for memory-saving reasons. 

Many works focusing on removing parameters from over-parametrized models~\cite{han2015learning,frankle2018the,tartaglione2018learning,sanh2020movement,marouf2024mini} propose simple magnitude-based schemes (which can naively be seen as norm ranking schemes to determine which are the least relevant parameters). When training a model with regularization like $\ell_1$ or even $\ell_2$ it is empirically validated, in a multitude of setups, that parameters in excess are likely to have low magnitude, and simple schemes like norm ranking constitute a robust baseline~\cite{kohama2023single,gupta2024complexity}. 

However, it is also known that global pruning strategies are impossible to deploy in the wild, due to different norm scaling in DNN's layers~\cite{hoffer2018norm,zhang2022all}. Indeed, the parameter's average magnitude varies layer-by-layer, depending on a multitude of factors that include initialization~\cite{glorot2010understanding,he2015delving} and the average backpropagation signal received~\cite{he2016deep} to name a few. One fair strategy would consist of ranking the adapters by their number of parameters (which would be the $\ell_0$ norm), motivated by the more significant potential expressivity of these. 

We argue that both the average parameter's norm and the adapter's cardinality (intended as the number of its parameters) should be accounted for, and we propose a metric estimating the average magnitude-per-parameter
\begin{equation}
    Z = \frac{{\lVert w_{f_n} \rVert}^2}{|w_{f_n}|},
    \label{eq:effada}
\end{equation}
where $Z$ is the relative score for the current layer, $\lVert\cdot\rVert$ is the $\ell_2$ norm, and $|\cdot|$ is the parameter's cardinality. We expect the larger the value in \eqref{eq:effada}, the higher the average norm per parameter, meaning that the current adapter is relevantly contributing to correcting the layer's output and should be considered ``important''. On the contrary, a low value of $Z$ indicates little perturbation of the layer's output and can be considered insignificant.

We exclude higher-order evaluation schemes for two reasons. The first one is that such an evaluation will be performed post-hoc, after training, meaning that we can expect first-order derivatives to be approximately zero, making first-order evaluation methods like~\cite{lee2018snip,tartaglione2022loss} not usable. In such context, higher-order methods should be used, which, however, are computationally expensive: to maintain their complexity at bay, approximations are introduced, making them unreliable in high-dimensionality~\cite{lecun1989optimal,yao2020pyhessian,yu2022hessian,yang2023global}. The second reason is that some of these methods introduce extra regularization constraints to make the model's pruning compliant with the pruning scheme~\cite{tartaglione2022loss,yang2023global}. At the same time, we want to remain as agnostic as possible to the training scheme, requiring simple $\ell_1$ or $\ell_2$ regularization schemes.\\
By quantifying the significance of adapter modules through $Z$ we can establish a ranking among them. This enables a trade-off strategy in which only the most relevant adapters are kept, while the least significant ones are discarded. Such a selection process requires a little cost in performance, as the removed adapters contribute little to the model’s output, while simultaneously reducing the overall memory footprint of the adapter set.

%% file: 4_experiments.tex
\section{Experiments}
\label{sec:experiments}
In this section, we provide details about the baseline architecture, the adaptation procedures explored in new domains, and the experiments we conducted with the adapter modules and memory-saving strategies, reporting the results we obtained.

\subsection{Setup}\label{sec:data}

\textbf{Data.} The studied datasets are three, provided by NASA's open-source database. The first one, called \textit{Synthetic Moon} dataset (shortly \textit{SMo}), contains $9.766$ realistic artificial renders of the lunar landscape, together with 36 real photos of the Moon's environment, \textit{Real Moon} or \textit{RMo},  (see \cite{smob} for more details about this dataset). The other two regarding Mars are named \textit{AI4Mars} \cite{AI4Mars} and \textit{MarsData-V2} \cite{marsdata} and consist of real photographs of the Mars landscape, $18.1$k and 835 images respectively.
We classified these data into three common classes: rocks, to be avoided by the rovers, safe terrain, to move upon, and finally a label identifying the sky. Further details on the dataset and the data preprocessing can be found in the Supplementary Material.

\noindent\textbf{Architecture.}
The chosen base architecture is U-Net~\cite{unet} due to its well-known high-performing feature extractor modules, the versatility of segmentation tasks, and the reduced training time. We performed a grid search to determine the best data augmentation techniques, loss function, and optimizer. We found the best solution for performing data augmentation with Random Crop (p = 0.50\%) on training images. All architectures have been trained using a Balanced Categorical Cross Entropy loss (with weighted classes).
The chosen optimizer was \textit{Adam} with a starting learning rate of 0.001. The learning rate was then reduced by a factor of 10 every time the validation loss didn't improve for 10 epochs until it reached $10^{-6}$.\\
We also tested different encoders, from the typical ResNet family \cite{resnet}, the traditional Vgg (version with batch norm layers \cite{vgg}), and also a version of the EfficientNet \cite{efficientnet}. Furthermore, we explored two light architectures devoted to smaller devices, like the MobileNetv2 \cite{mobilenet2} and the lightest version of MobileOne \cite{mobileone}, discarding more complex ones like the DenseNets \cite{densenet}, due to their higher computational cost, making them unsuitable for space applications.\\
For the same reasons, although their elevated performance, we didn't study the Transformer~\cite{segformer} architecture, because of its computational costs and memory size.\\
All the implementation details are provided in the Supplementary Material.

\subsection{Main Results}
\label{sec:adapt_domains}
\begin{table*}[t]
\centering
\caption{U-Net encoders study on the Synthetic Moon dataset.}
\label{tab:encoders}
\begin{adjustbox}{max width=0.8\textwidth}
\begin{tabular}{ccccccccc}
\toprule
\multirow{3}{*}{\bf Encoder} & \textbf{Total} & \multirow{2}{*}{\bf FLOPs} & \multicolumn{2}{c}{\textbf{Balanced}} & \multicolumn{2}{c}{\multirow{2}{*}{\bf IoU [\%]}} & \multirow{2}{1.8cm}{\centering\bf Storage\\ Memory} & \multirow{2}{1.8cm}{\centering\bf Compressed \\(.tar.gz) }\\
 & \textbf{Params} &  & \multicolumn{2}{c}{\textbf{Accuracy [\%]}}  &  &   &   & \\
 & \textbf{[M]} & \textbf{[G]} & Validation & Test & Validation & Test  & \textbf{[MB]} & \textbf{[MB]} \\
 \midrule
ResNet-18 & \bf14.32 & \bf21.42 & \bf96.49 & \bf94.98 & \bf83.76 & \bf82.32  & \bf54.75 & \bf50.75 \\
ResNet-34 & 24.43 & 31.12 & 96.46 & 94.89 & 82.93 & 82.05 & 93.37 & 86.44 \\
ResNet-50 & 32.52 & 42.58 & 96.35 & 94.96 & 83.37 & 81.99 & 124.39 & 115.20 \\
\midrule
EfficientNet-b3 & 13.16 & 15.75 & 96.16 & 94.77 & 85.28 & 83.10 & 50.79 & 47.08\\
MobileOne-s0 & 8.59 & 18.95 & 90.99 & 92.40 & 76.95 & 79.35 & 33.62 & 30.80\\
MobileNet-v2 & 6.63 & 13.69 & 94.89 & 93.45 & 77.43 & 77.63 & 25.57 & 23.61\\
\midrule
Vgg11-BN & 18.26 & 57.94 & 96.09 & 94.73 & 82.57 & 81.29 & 69.74 & 64.60\\
Vgg13-BN & 18.44 & 77.39 & 96.16 & 94.76 & 82.68 & 81.24 & 70.45 & 65.35\\
Vgg16-BN & 23.76 & 99.17 & 96.46 & 95.00 & 83.28 & 82.11 & 90.73 & 84.21\\
Vgg19-BN & \bf29.07 & \bf120.94 & \bf96.58 & \bf95.16 & \bf83.48 & \bf82.18 & \bf111.01 & \bf103.05 \\
\bottomrule
\end{tabular}
\end{adjustbox}
\end{table*}

\begin{table*}[t]
\centering
\caption{Performance on adapted domains: Moon and Mars.}
\begin{adjustbox}{max width=\textwidth}
\begin{tabular}{ccccc|cccc}
\toprule
\multirow{ 2}{*}{\bf Method} & \multirow{ 2}{*}{\bf Encoder} & \bf Trained & \bf Trained &\bf Forward  & \multicolumn{4}{c}{\bf Balanced Accuracy (\%)}\\
    &   &  \bf Params (M) &  \bf Params (\%) &  \bf FLOPs (G) &   \textit{RMo} & \textit{AI4Mars} & \textit{Marsdataset} & Mean \\
\midrule
Baseline & ResNet-18 & 14.32 & 100.00 & 21.42 & 82.14 & 34.05 & 57.61 & 57.93\\
Scratch & ResNet-18 & 14.32 & 100.00 & 21.42 & 85.41 & 87.89 & 86.96 & 86.75\\
Full finetuning & ResNet-18 & 14.32 & 100.00 & 21.42 & \bf 92.70 & 90.53 & \bf93.46 & \bf92.23 \\
Encoder finetuning & ResNet-18 & 11.17 & 77.99 & 21.42 & 90.84 & \bf 91.80 & 92.36 & 91.67\\
Decoder finetuning & ResNet-18 & 3.15 & 22.01 & 21.42 & 90.03 & 86.35 & \bf 93.46 & 89.95 \\
Batchnorm finetuning & ResNet-18 & 0.01 & \textcolor{blue}{0.007} & 21.42 & 85.44 & 83.00 & 88.17 & 85.54 \\
\midrule
Adapters (All) & ResNet-18 & 1.58 & 9.94 & 23.57 & 89.68 & 89.99 & 91.53 &90.82 \\
Adapters (Ranked and fused) & ResNet-18 & \textcolor{blue}{0.13} & \textcolor{blue}{0.82} & 21.27 & 89.48 & 80.30 & 90.61 & 86.80 \\
\midrule
\midrule
Baseline & Vgg19-BN & 29.07 & 100.00 & 120.94 & 70.08 & 35.21 & 53.54 & 52.94 \\
Scratch & Vgg19-BN & 29.07 & 100.00 & 120.94 & 82.76 & 86.71 & 83.92 & 84.46 \\
Full finetuning & Vgg19-BN & 29.07 & 100.00 & 120.94 & \bf 93.65 & 90.34 & 91.64 & 91.88 \\
Encoder finetuning & Vgg19-BN & 20.03 & 68.92 & 120.94 & 90.48 & \bf 91.93 & \bf 93.41 & \bf 91.94 \\
Decoder finetuning & Vgg19-BN & 9.03 & 31.08 & 120.94 & 89.14 & 85.97 & 89.30 & 88.14 \\
Batchnorm finetuning & Vgg19-BN & 0.02 & \textcolor{blue}{0.007} & 120.94 & 88.43 & 81.26 & 90.98 & 86.89 \\
\midrule
Adapters (All) & Vgg19-BN & 3.11 & 9.68 & 136.18 & 90.10 & 90.69 & 87.96 & 89.38 \\
Adapters (Ranked and fused) & Vgg19-BN & \textcolor{blue}{0.48} & \textcolor{blue}{1.49} & 120.47 & 89.36 & 86.18 & 87.81 & 87.78 \\
\bottomrule
\end{tabular}
\end{adjustbox}
\label{tab:finetuning_adapters_domains}
\end{table*}

In our experiments, we include the following strategies.
\begin{itemize}[noitemsep]
    \item \textit{Baseline}: use the pre-trained neural network on \textit{SMo} to predict class labels on adaptation domains.  
    \item \textit{Scratch}: training from scratch a new neural network on adaptation domains. 
    \item \textit{Full finetuning}: fine-tuning both the encoder and decoder of the baseline neural network.
    \item \textit{Encoder finetuning}: fine-tuning just the encoder.
    \item \textit{Decoder finetuning}: fine-tuning just the decoder.
    \item \textit{Batchnorm finetuning}: training batchnorm layers and keeping everything else frozen. 
    \item \textit{Adapters (All)}: training adapter modules and freezing the rest of the pre-trained neural network.
    \item \textit{Adapters (Ranked and fused)}: selecting the most important adapters with the ranking method and fuse them in the original architecture.
\end{itemize}
\begin{figure*}[t]
    \centering
    \includegraphics[width=\textwidth]{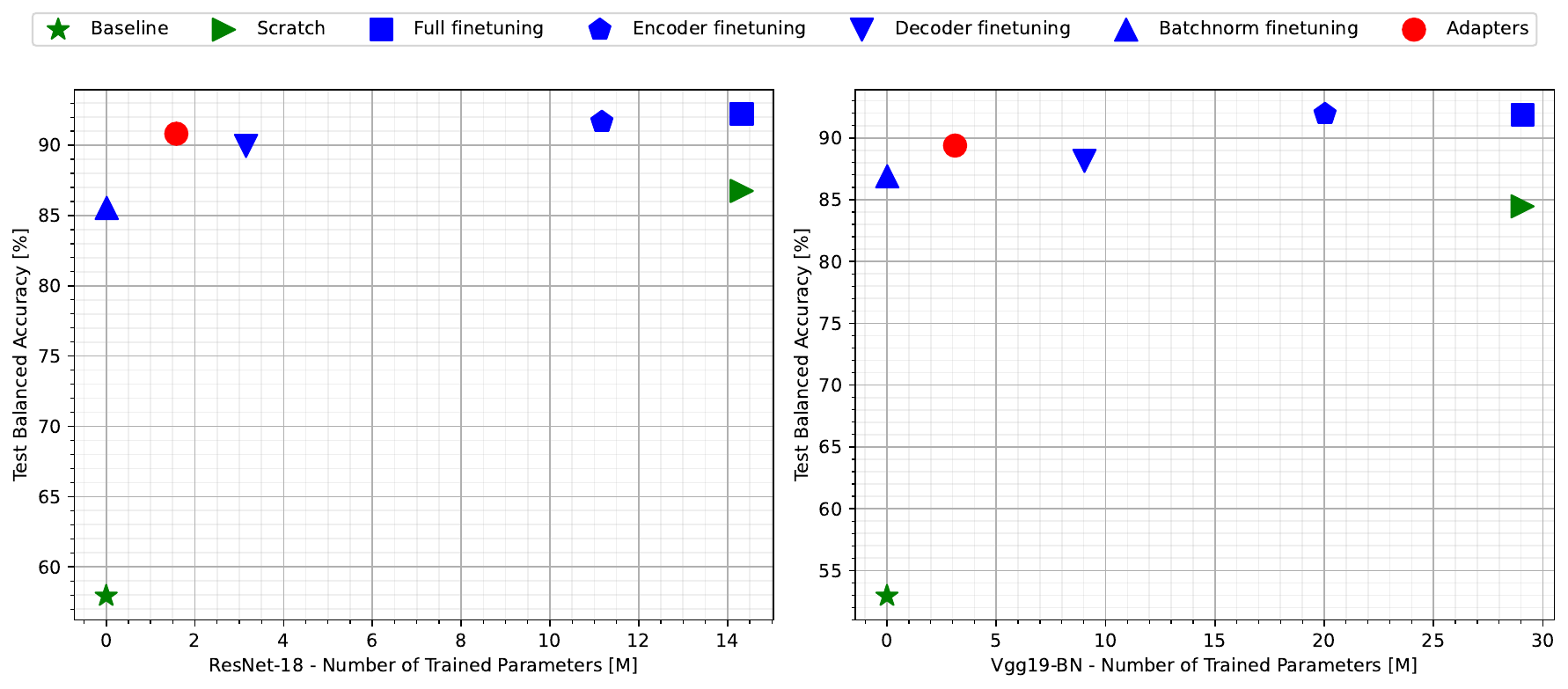}
    \caption{Pareto curves for VGG19-BN and ResNet-18.}
    \label{fig:pareto}
\end{figure*}
\begin{figure*}[t]
    \centering
    \includegraphics[width = 0.95\textwidth]{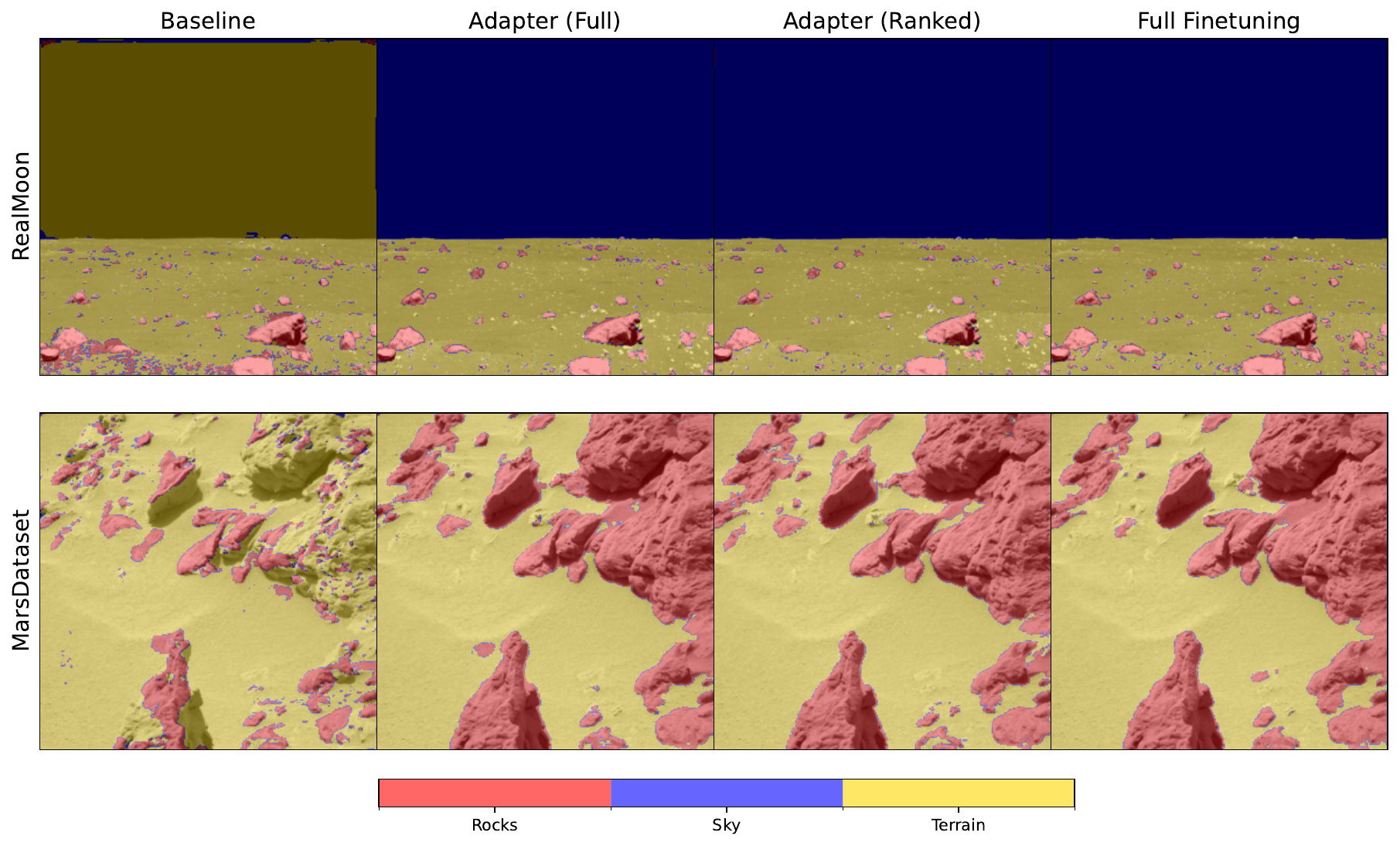}
    \caption{Comparison of U-Net predictions on real Moon and real Mars images using different methods.}
    \label{fig:realdomain}
\end{figure*}

The first way consisted of the study of different grades of re-training: a complete one from scratch, maintaining just the same architecture (called \textit{Baseline}), a complete finetuning of the synthetic lunar model, adjusting the weights on the new domain (the \textit{Full finetuning} model) and then the finetuning of just some layers of the initial model (\textit{Encoder}, \textit{Decoder} and \textit{Batchnorm}). On the other hand, by introducing adapters, we studied the case of training just the adapters on the new domain while freezing the network on the initial synthetic lunar weights. Furthermore, the measurements of the application of the new strategies are reported too; their implementation will be detailed in the following sections. 
Together with the performance differences between the algorithms, the p-values are also provided. They are evaluated using a paired t-test or a Wilcoxon test to determine whether the values are normally distributed. The difference is statistically significant if $p<0.05$.
The results reported in Table~\ref{tab:finetuning_adapters_domains} show not only the performances of these different adaptation strategies together with the FLOPs counting but also the trained parameters (and their ratio on the total ones) as a comparison. It can be observed that the adapter method has an extreme gain in the number of trained parameters with a relatively small increase in the number of operations (FLOPs) and slightly worse performance. At the cost of $\approx10\%$ of trained parameters, a similar level of performance of a completely re-tuned model can be achieved.\\
This behavior can also be visualized in Pareto curves (Fig.~\ref{fig:pareto}) showing the mean balanced accuracy versus the number of trained parameters. As can be seen, the top left corner of the plots corresponds to adapters, meaning that high performance can be reached with a relatively small training price. 
Notice that the architecture of reported adapters in Table~\ref{tab:finetuning_adapters_domains} is a module composed of a BN layer connected to a convolutional $1\times1$ one. Different structures were studied and their performances are presented in Sec.~\emph{Adapters architecture}.\\
Considering each domain specifically, the best choice for real lunar images is to keep the pre-trained weights from the synthetic lunar model and fine-tune them to adjust to the new domain. The adapter application helps with the memory issue at the cost of a slightly worse performance($\approx3\%$, statistically significant $p=0.02$).
The same considerations can be made for the Martian domains, especially about fine-tuning. Due to the obvious differences between the lunar and Martian environments, it's not mandatory to keep the already performed training. 
On the other hand, the adapter introduction brings the same advantages as before, reaching even nearer levels of performance to the more complete retrained models. In particular, the AI4Mars's adapters implemented in the Vgg19-BN lose just $\approx 1.2\%$ ($p=1.1e-5$) more than the best method, i.e., the encoder finetuning one. In contrast, the MarsDataset ones perform better with ResNet-18 encoder, worsening the full finetuning performance by only $\approx 1.9\%$ ($p=2.3e-5$). A prediction example with a comparison between adaptation methods on the MarsDataset and RealMoon domains can be seen in Fig.~\ref{fig:realdomain}. \\
A preliminary noise robustness study, where the baseline model and the adapter-equipped one's performance are compared under different noise conditions, is presented in the Supplementary Material.

\subsection{Adapters Architecture}
\label{sec:adapters_architecture}
\begin{wraptable}{r}{0.5\textwidth}
\small
\vspace{-12pt}
\centering
\caption{Ablation on different adapter designs on \textit{RMo}.}
\label{tab:adapterdesigns}
\begin{tabular}{ccccc}
\toprule
\multicolumn{4}{c}{\bf Adapter design}&\multirow{2}{*}{\bf Balanced Accuracy (\%)}\\
 BN & ReLU & Conv1x1 & BN &  \\
\midrule
\bf \checkmark &        & \bf \checkmark &         & \bf 91.42 \\ 
\checkmark & \checkmark & \checkmark &         & 89.66 \\ 
         &        & \checkmark & \checkmark & 75.84 \\ 
\bottomrule
\end{tabular}
\end{wraptable}
We compared here different adapter configurations, visualized in the Supplementary Material. 
In Table~\ref{tab:adapterdesigns}, the results of their performance on the RMo dataset (validation set) are presented. Interestingly, introducing a non-linearity such as a ReLU function didn't bring any improvement, validating that the best design results in the sequence of a BN layer followed by a convolutional $1\times1$ one. We infer this is due to the tight relationship between domains, where simple linear corrections at the layer's scale are sufficient to improve performance dramatically. Employing higher complexity makes the optimization task harder, meaning that finding the optimal training policy is much harder - this justifies the performance drop in those cases.

\subsection{Adapters Ranking}
\label{sec:adapter_ranks_esec:xperiments}
\begin{figure*}[t]
\centering
\includegraphics[width = \textwidth]{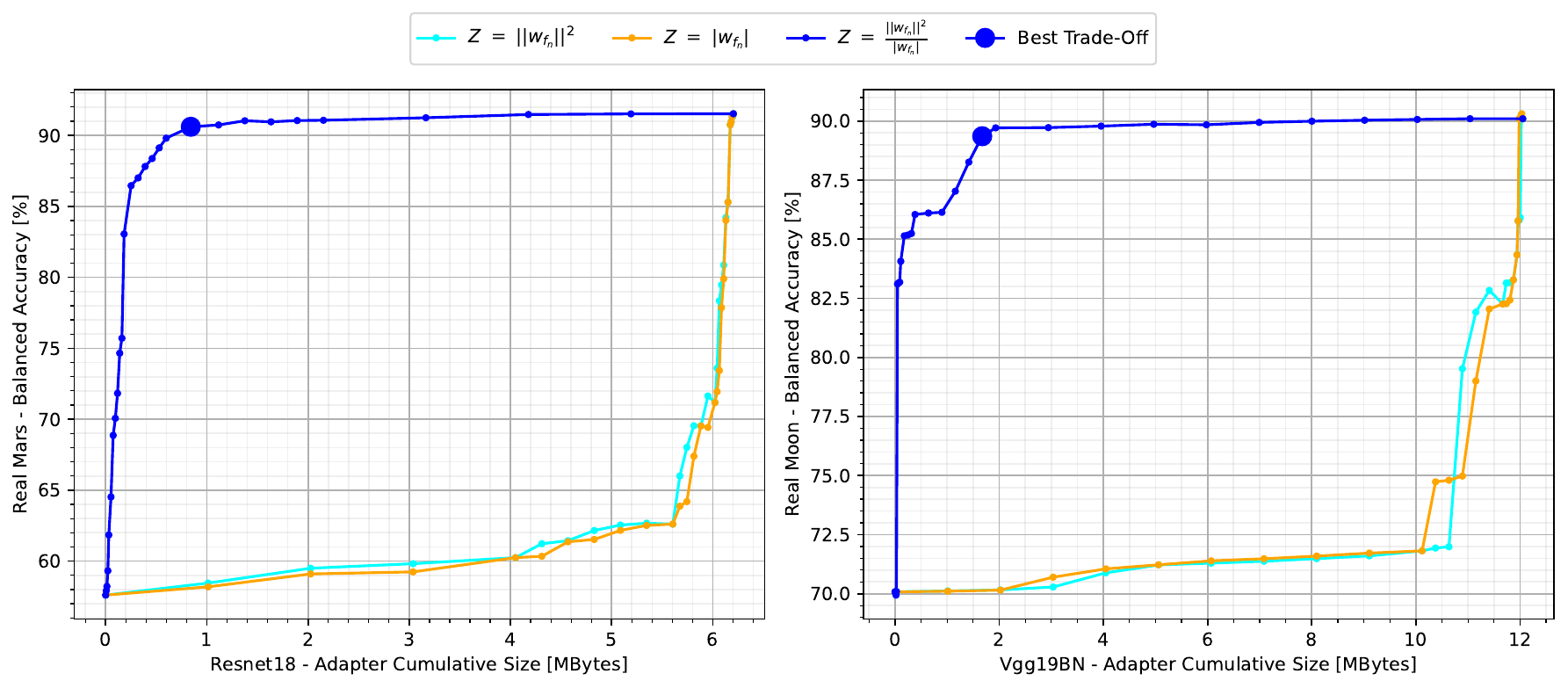}
\caption{Performance as a function of the cumulative size of inserted adapters: left and right images show respectively the results on real mars (MarsDataset) and Real Moon domains (test sets).}
\label{fig:adapters_rank}
\end{figure*}

Subsequently to the adapter introduction, as described in Sec.~\ref{sec:adapt_ranks}, we perform an adapter evaluation, ranking them from the most significant to the least one through different scores. The Pareto curves (Fig.~\ref{fig:adapters_rank}) show the balanced accuracy behaviour as a function of the cumulative size of the selected adapter up to the total adapter set. From them, it's clearly visible that the third score, the square norm normalized by the layer size (see Sec.~\ref{sec:adapt_ranks}), represented in blue, provides the best adapters ranking, reaching first the best performance at the smaller cumulative size.
To select the best trade-off in the adapter removal, we accept a worsening of $0.5\%$ from the top performance (validation-wise), cutting away the other less significant adapters and saving memory.
For example, for MarsDataset we noticed that only 18 adapters out of the 27 total adapters are truly necessary for the Resnet-18 U-Net. By using only 18 adapters we have a slight decrease in performance ($\approx -1.0 \% \ \text{max}$), but a huge decrease in memory consumption: instead of all of the adapter set, in terms of parameters, just the $8.2\%$ (Resnet-18) and $15.4\%$ (Vgg19-BN) of them can be used (see Table~\ref{tab:finetuning_adapters_domains}).\\
The position of the removed modules is also quite interesting: these turn out to be the heaviest ones (size-wise) and, due to the U-Net architecture, they are placed in the deepest part of the architecture. This means that the deeper the block, the less important it is, implying that the most important adapters for feature extraction and classification are in the first encoder layers and the last decoder layers. A scheme of these can be seen in Fig.~\ref{fig:resnet18_ranking} (see Supplementary Material for Vgg).
Notice that to perform this strategy, it is still necessary to train all adapters and then proceed with measuring their ranks and, subsequently, the selection. For this reason, in Table~\ref{tab:finetuning_adapters_domains} the trained parameters are presented in double mode: all of the adapter parameters are effectively trained, and only then the ranking pruning is applied, massively cutting their numbers.\\
In the sense of a practical application, these results mean that we can reduce the update memory size for a different domain using the adapter modules, and we can also further save memory by sending just a fraction of the total number of adapters, the most significant ones, with minimal performance loss.

\begin{figure*}[ht]
    \centering
    \includegraphics[width = \textwidth]{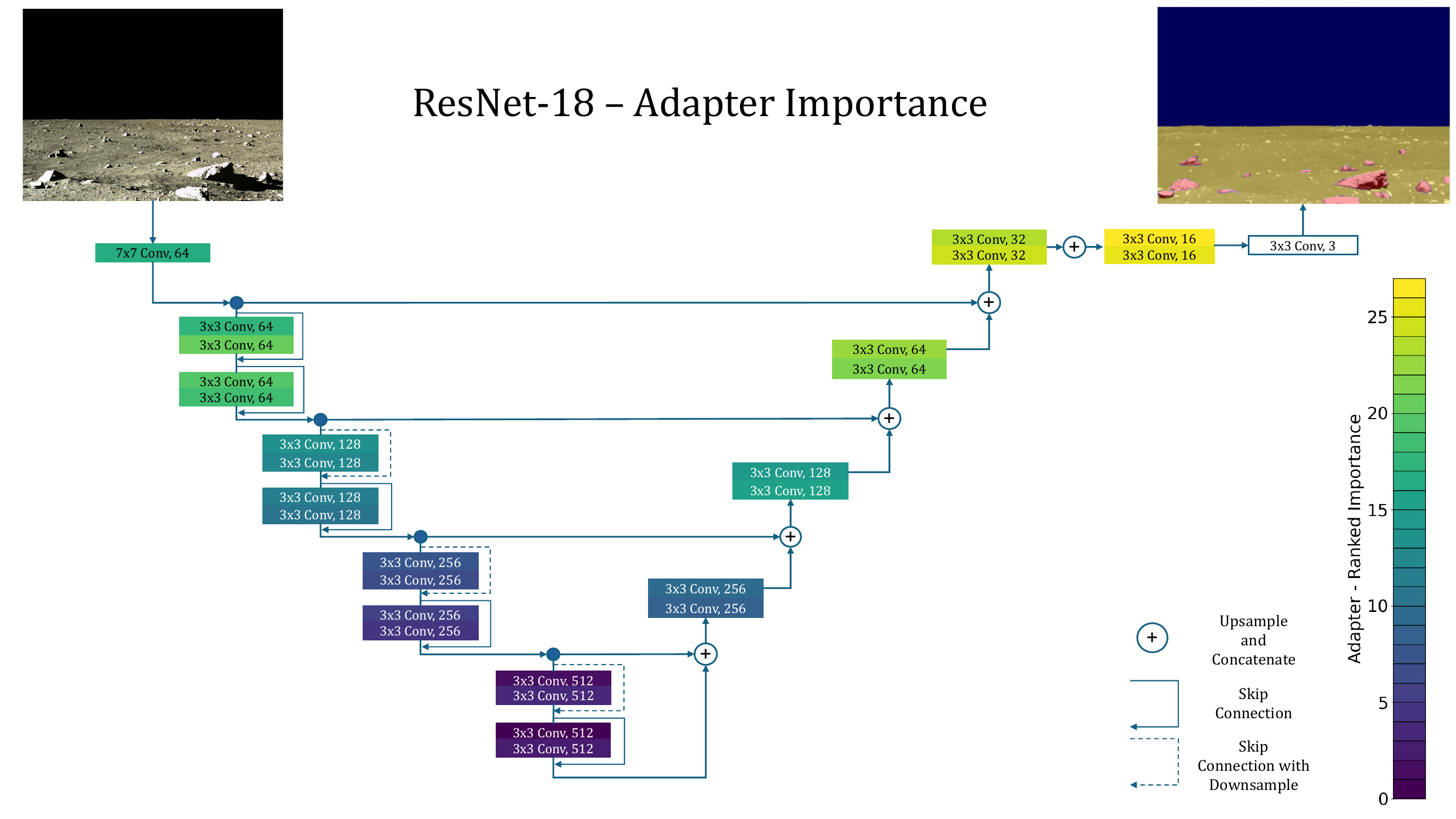}
    \caption{Visualization of Resnet-18 adapter ranking: modules near the bottleneck of the U-Net and with larger size are less important than the ones in the first encoder layers and last decoder layers}
    \label{fig:resnet18_ranking}
\end{figure*}

\subsection{Adapters Fusion}
\label{sec:adapter_fusion_experiments}

In the end, employing the adapters fusion strategy thoroughly described in Sec.~\ref{sec:adapter_fusion}, we produce a ``fused'' version of the previously adapted models, absorbing the ranked adapters inside the baseline architecture. Applying this process also decreases the computational number of operations (i.e. FLOPs) while keeping the same level of performance (average $p=0.59$, differences extremely not statistically significant). \\
The ``fusion" procedure doesn't affect the performance. Still, the main gain comes in combination with the ranking one: while the ranking strategy prunes the number of required adapters, saving a huge amount of memory, the fusion one simplifies the architecture, decreasing the overall FLOPs and zeroing the cost of the adapters introduction. Furthermore, the "fused" model now has the same structure as the initial one, making other updates, modifications, or module insertions (like other future adapters) more feasible.\\
The values can be seen in Table~\ref{tab:finetuning_adapters_domains}: while just the ranking strategy worsens the performance, the model size and the FLOPs number return to the baseline models' level, decreasing both of them by $\approx10\%$ than utilizing all of the adapter set. Notice also that the storage memory of the fused architecture reaches the same level as the baseline one (see also Table~\ref{tab:encoders}), proving that in the end, we will have nearly similar architectures, but now with the changed adapted parameters.

\subsection{Inference on Embedded Devices} \label{sec:devices}
As previously shown, the main advantage of the adapter's implementation is the flexibility to adapt to a new domain, keeping the same level of performance. Memory is one major factor to consider for embedded devices, due to their limited resources. The hardware cannot handle a complete re-training by itself, so a lighter approach, such as a small update from Earth, can keep the performance on level. This gain might be negligible if the computational training power can manage a completely new training, but often the on-field devices are not able to do so, performing just the inference phase of the trained model.\\
Due to these limitations, it's fundamental to measure the impact of the adapter on the model, especially regarding the inference time and the memory usage, while loaded on an embedded device. To simulate an on-the-field situation, we decided to test two different hardware: a Raspberry Pi 4 board, equipped only with a CPU processor, and a Jetson Orin Nano one, with a Nvidia GPU, measuring the inference time, the memory usage, and FLOPs numbers while working on board. This choice allowed us to compare the different performances between the two hardware (CPU - GPU) of our models. It is worth noting that, although GPUs have not been implemented yet on on-site rovers, the increase in the onboard computing performance demands might accelerate the usage of such devices in space~\cite{gpuspace}, which is the reason why we also included a GPU device in our inference experiments. Further details on the two devices can be found in the Supplementary Material. \\
A comprehensive collection of the performances of the previously presented models is also reported in the Supplementary Material, while Table~\ref{tab:final_comparison} reports a summary of performance and computational effort (in terms of FLOPs and inference time). The adapter usage worsens the performance and increases the total model size and complexity, but the update cost is drastically reduced. If the two new proposed strategies are performed, even these costs are zeroed out: memory footprint, storage size, and computational cost (i.e., FLOPs and inference time) are returned to baseline level, at the cost of even less updated memory dimension.

%% file: 5_discussion.tex
\section{Discussion}

\begin{table}[t]
\centering
\caption{Final memory and computational cost comparison. "None" means the finetuning of the baseline model without the adapters. "Ranked", "Fused", and "Energy costs" values are dataset-specific (ResNet-18 - MarsDataset, Vgg19-BN - RMo).}
\label{tab:final_comparison}
\small
\begin{adjustbox}{max width=\textwidth}
\begin{tabular}{cccccccccccc}
\toprule
\multirow{3}{1cm}{\bf Encoder} & \multirow{3}{1cm}{\centering\bf Adapter\\ type} & \multirow{3}{1.8cm}{\centering \bf Balanced \\ Accuracy [\%]} & \multicolumn{2}{c}{\multirow{2}{2cm}{\centering\bf FLOPs [G]}} & \multicolumn{2}{c}{\multirow{2}{2cm}{\centering\bf Storage Memory [MB]}} & \multicolumn{2}{c}{\multirow{2}{2cm}{\centering\bf Compressed (.tar.gz) [MB]}} & \multirow{3}{1.8cm}{\centering\bf Energy\\ Spent [kJ]} & \multicolumn{2}{c}{\multirow{2}{*}{\centering\bf Inference [s]}}  \\
 & & & & & & & & &  \\
 & & & \textit{Total} & \textit{Update} & \textit{Total} & \textit{Update} & \textit{Total} & \textit{Update} & & RPi4 & JNano \\
 \midrule
\multirow{4}{*}{\rotatebox[origin=c]{90}{ResNet-18}} & None & \bf93.46 & 21.42 & +21.42 & 54.79 & +54.79 & 50.78 & +50.78 & 39.36 & 6.96 & 0.60 \\
 & All & 91.53 & 23.57 & +2.15 & 60.91 & +6.12 & 56.42 & +5.64 &27.28 & 8.29 & 0.77 \\
 & Ranked & 90.61 & 22.96 &  +1.54 & 55.61 & +0.82 & 51.54 & +0.76  & //  & 7.51 & 0.69\\
 & Fused & 90.61 & \bf21.27 & \bf-0.15 & \bf54.67 & \bf-0.08 & \bf50.80 & \bf+0.02 &// & \bf7.05 & \bf0.63\\
\midrule
\multirow{4}{*}{\rotatebox[origin=c]{90}{Vgg-19-BN}} & None & \bf93.65 & 120.94  & +120.94 & 111.01 & +111.01 & 55.00 & +55.00 & 2.37 & 25.78  & 0.95 \\
 & All & 90.10 & 136.18 & +15.24 & 123.00 & +11.99 & 66.04 &+11.04 & 2.36 & 33.26  & 1.45 \\
 & Ranked & 89.36 & 131.79 & +10.85 & 112.72 &+1.71 & 56.57 &+1.57  & // & 28.12  & 1.11 \\
 & Fused & 89.36 & \bf120.47  & \bf-0.47 & \bf110.88 & \bf-0.13 & \bf103.03 & \bf+48.03 & // & \bf25.86  & \bf0.97 \\
\bottomrule
\end{tabular}
\end{adjustbox}
\end{table}

In Table~\ref{tab:final_comparison} we reported the final comparison between the presented steps, following also a hypothetical order of procedure. A hypothetical application is a situation of an on-field device equipped with a DL model for rock segmentation (for example, for autonomous driving). Given such devices' computational and hardware constraints, both the baseline training and subsequent model updates must be performed at ground stations before being transmitted to the deployed system. \\
For instance, in a lunar environment, the baseline is trained over a synthetic dataset providing a set of generic features (e.g. \textit{SMo}). Then, we make the case that upgrades are needed because of the change in the environmental conditions: the model should be adapted to a new domain that is similar but at the same time different from the baseline starting one. Note that this adjustment is possible only if a large enough dataset of the new environment is already provided. \\
First of all, the simplest solution consists of a completely new finetuning and adjustment of the parameters to the new domain. The \textit{None} lines show the cost of a complete finetuning in terms of FLOPs and Storage Memory, considering re-training $100\%$ of the model without the use of the adapter modules. Doing so requires re-sending the full-finetuned model entirely to the devices (see Storage Memory update size equals to the total one), making difficult the update due to the bandwidth constraints, for example.\\
Instead, we presented the advantages of the adapter's usage in this work. As a first approach, an entire set of the trained-on-new-domain adapters can be used to update the model (namely, the \textit{All} option). While this approach marginally increases computational costs (e.g., $+2.15$~GFLOPs), it significantly reduces the memory required for updates (e.g., $+6.12$~MB). Additionally, training only the adapters requires less energy. \\
These benefits can be further improved: thanks to the "adapter ranking" strategy just a tiny subset of the adapter set can be sent (see \textit{Ranked}), reducing even more the memory cost (e.g. $+0.82$~MB) and limiting a little bit the computational operations (e.g. $+1.54$~GFLOPs). However, this reduction comes at the expense of a slight worsening of performance (see Sec.~\ref{sec:adapt_ranks}), it could provide a viable trade-off, matching the space exploration constraints.
Finally, these selected adapters can be absorbed within the architecture by the "adapter fusion" strategy (\textit{Fused}): this results in a model with computational and storage costs nearly identical to the original baseline, while still being adapted to the new domain with minimal performance loss. Importantly, the fusion operation can be performed directly on the device, meaning the effective update cost remains the same as the \textit{Ranked} configuration. However, the computational overhead introduced by adapters is eliminated, simplifying the model architecture and facilitating future updates. The memory saving on the update size using these strategies fits well with the usual bandwidth limitations in space exploration, together with a low computational cost, due to the space-certified hardware characteristics.

\section{Conclusion}

In this work, we studied the rock segmentation problem on real photos of the Moon and Mars exploiting the adapter modules with two different memory-saving strategies. We show that adapters have advantages over performing a complete re-training of the base architecture for each dataset regarding the space exploration constraints in bandwidth communication and flexibility. We obtain segmentation performances comparable to complete retrained models with adapters ($\approx\!10\%$ of the parameters). We also showed two techniques to exploit these advantages further. \emph{Adapters ranking} is the first, showing that just a smaller portion of adapters is genuinely significant, pruning some of them and saving more memory ($\approx86\%$), at a slight cost in performance ($\approx\!0.5\%$). Then, \emph{adapter fusion} fits them into the architecture and lowers the computational cost ($\approx\!10\%$). The memory saving is less significant ($<\!10\%$), but this strategy keeps the architecture simpler and faster than before, making other modifications more feasible. We are confident these findings will enable research in resource-constrained environments, where computation is critical and where bandwidth for updates is very limited.

\section{Data availability}
The data used in this article comes from publicly available resources. The datasets used during the current study are available for the \href{ https://www.kaggle.com/datasets/romainpessia/artificial-lunar-rocky-landscape-dataset}{Synthetic Moon dataset}, \href{https://data.nasa.gov/Space-Science/AI4MARS-A-Dataset-for-Terrain-Aware-Autonomous-Dri/cykx-2qix/about_data}{AI4Mars}, and \href{https://github.com/CVIR-Lab/MarsData}{Marsdataset}. LO and ESM re-annotated Marsdataset, where the new ground-truths can be found at \small{\url{https://github.com/lolivi/MoonMarsAdapters/tree/main/marsdataset-sky-annotations}}.

%% file: supplementary.tex
\onecolumn
\section*{Supplementary Material}



\input{suppl_data}


\begin{figure*}[ht]
    \centering
    \includegraphics[width = 0.76\textwidth]{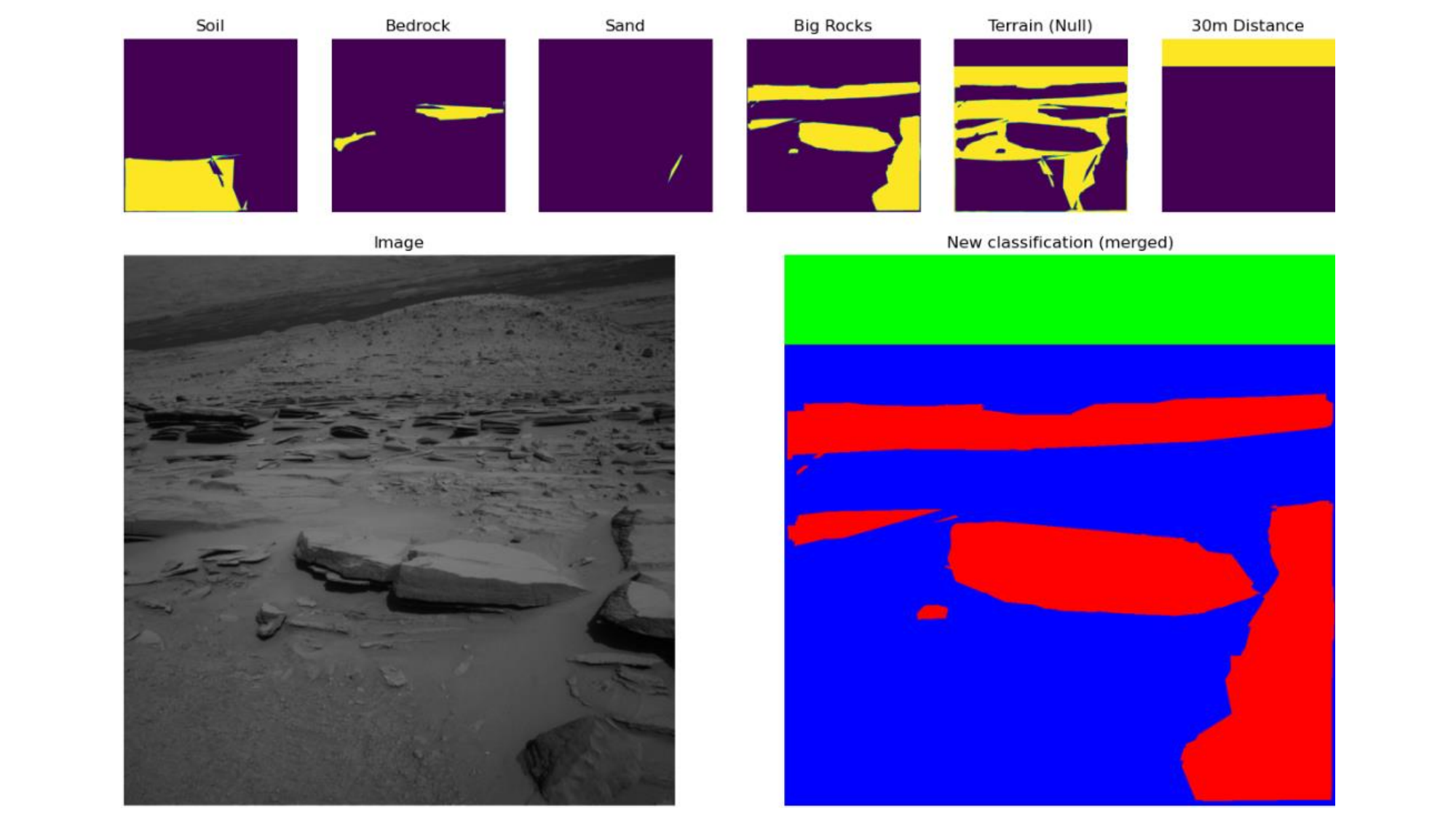}
    \caption{Example of merging class labels in AI4Mars dataset: blue, red, and green colors represent respectively terrain, rocks, and sky. Labeling is extremely rough and unreliable with respect to other domains.}
    \label{suppl_fig:mars-merging}
\end{figure*}

\begin{figure*}[ht]
    \centering
    \includegraphics[width = 0.76\textwidth]{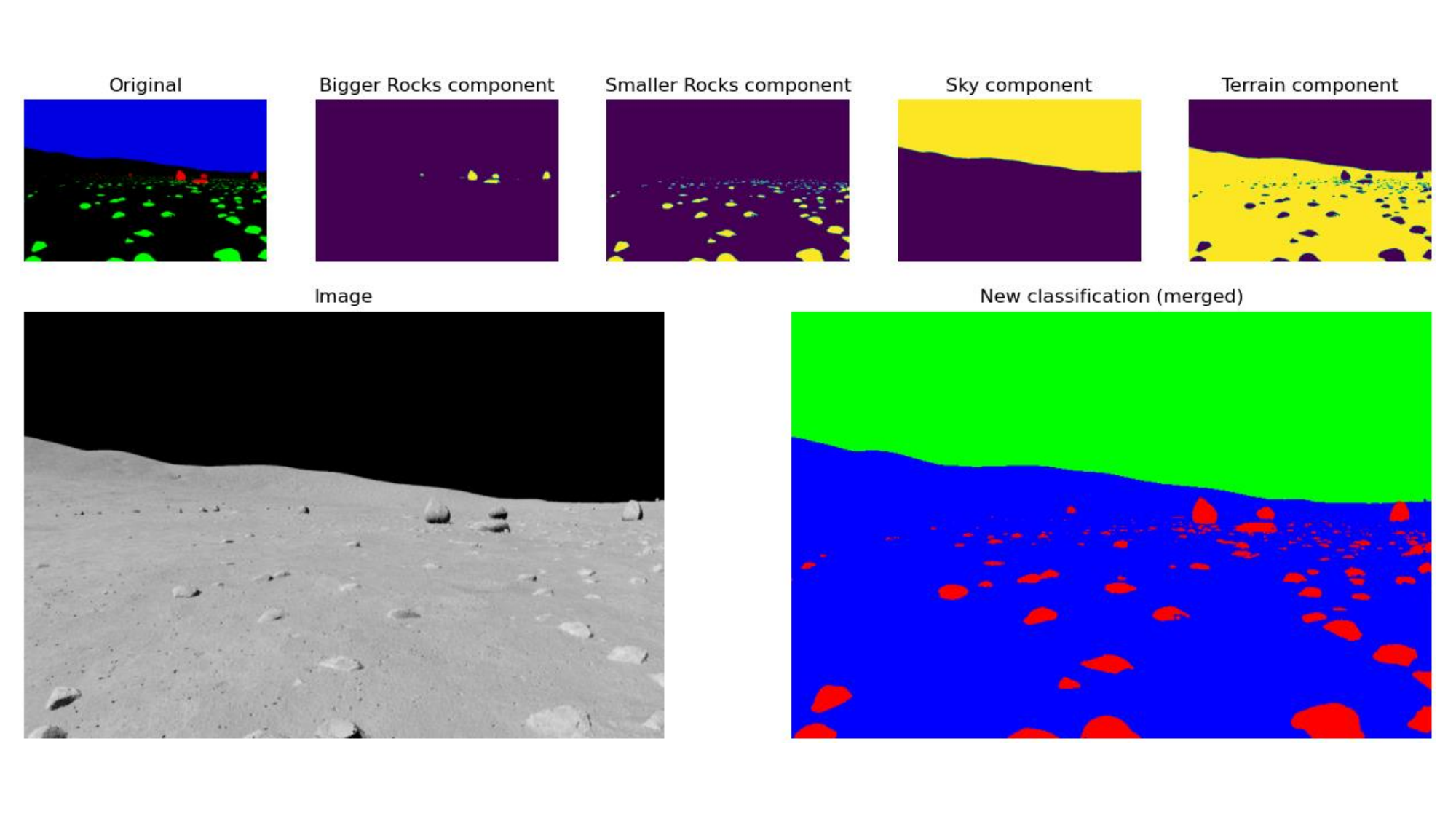}
    \caption{Example of merging class labels in Synthetic and Real Moon datasets: blue, red and green colours represent respectively terrain, rocks ,and sky.}
    \label{suppl_fig:moon-merging}
\end{figure*}

\newpage
\subsection{Adapter designs schemes}

\begin{figure*}[ht]
\begin{subfigure}[b]{0.45\textwidth}
     \centering
     \includegraphics[scale=0.35]{figures/Scheme_adapter_fusion-cropped-1.pdf}
     \caption{BN + Conv1x1 }
\end{subfigure}
\hspace{-0.5cm}
\begin{subfigure}[b]{0.45\textwidth}
     \centering
     \includegraphics[scale=0.35]{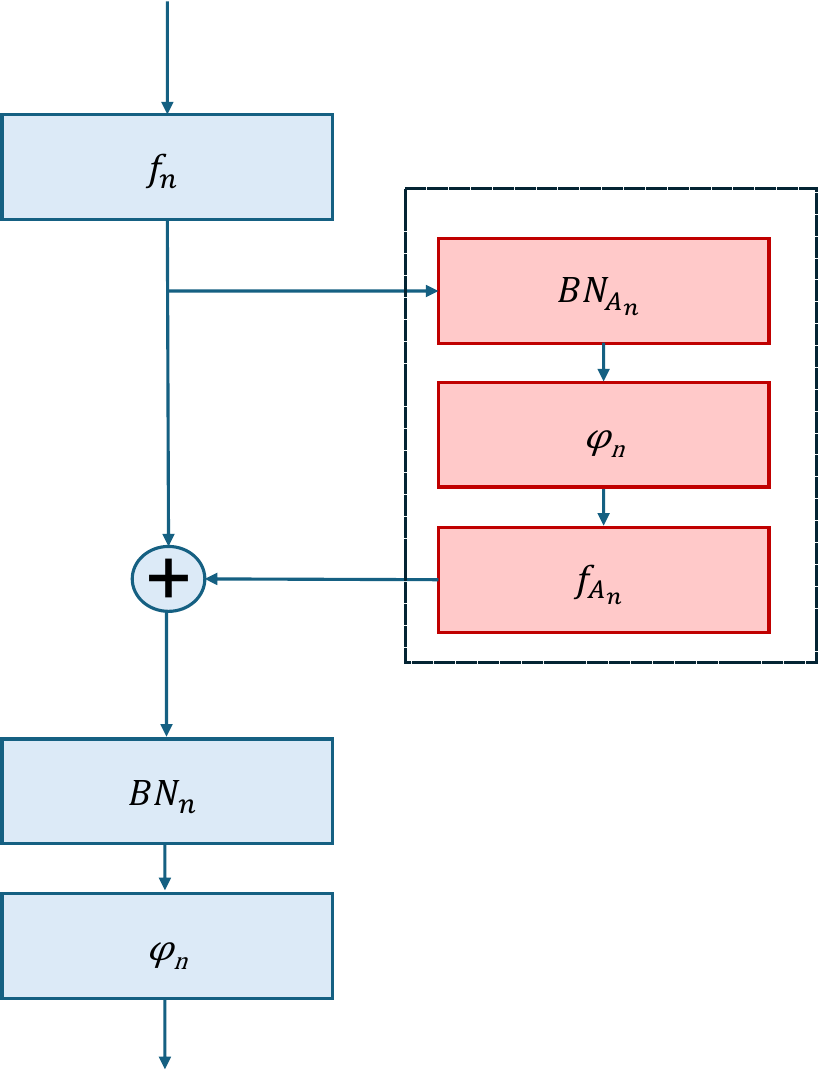}
     \caption{BN + ReLU + Conv1x1}
\end{subfigure}
\centering
\\
\begin{subfigure}[b]{0.45\textwidth}
     \centering
     \includegraphics[scale=0.35]{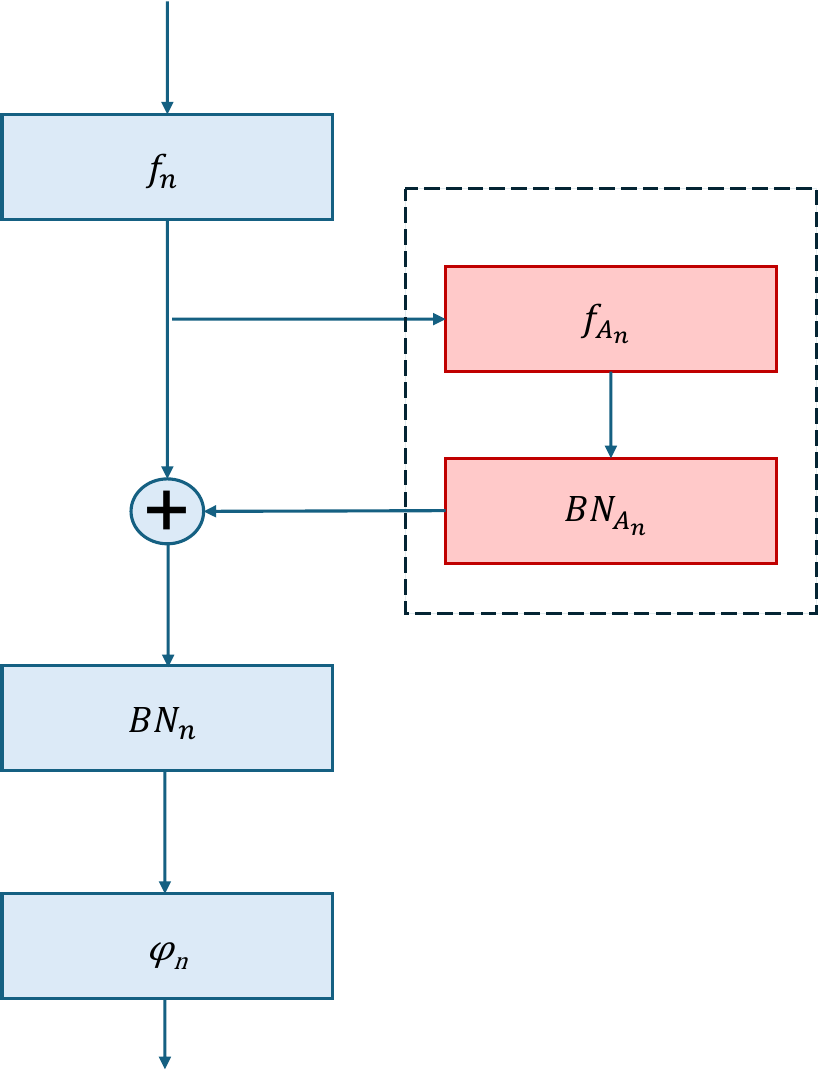}
     \caption{Conv1x1 + BN}
\end{subfigure}
\hspace{-0.5cm}
\begin{subfigure}[b]{0.45\textwidth}
     \centering
     \includegraphics[scale=0.35]{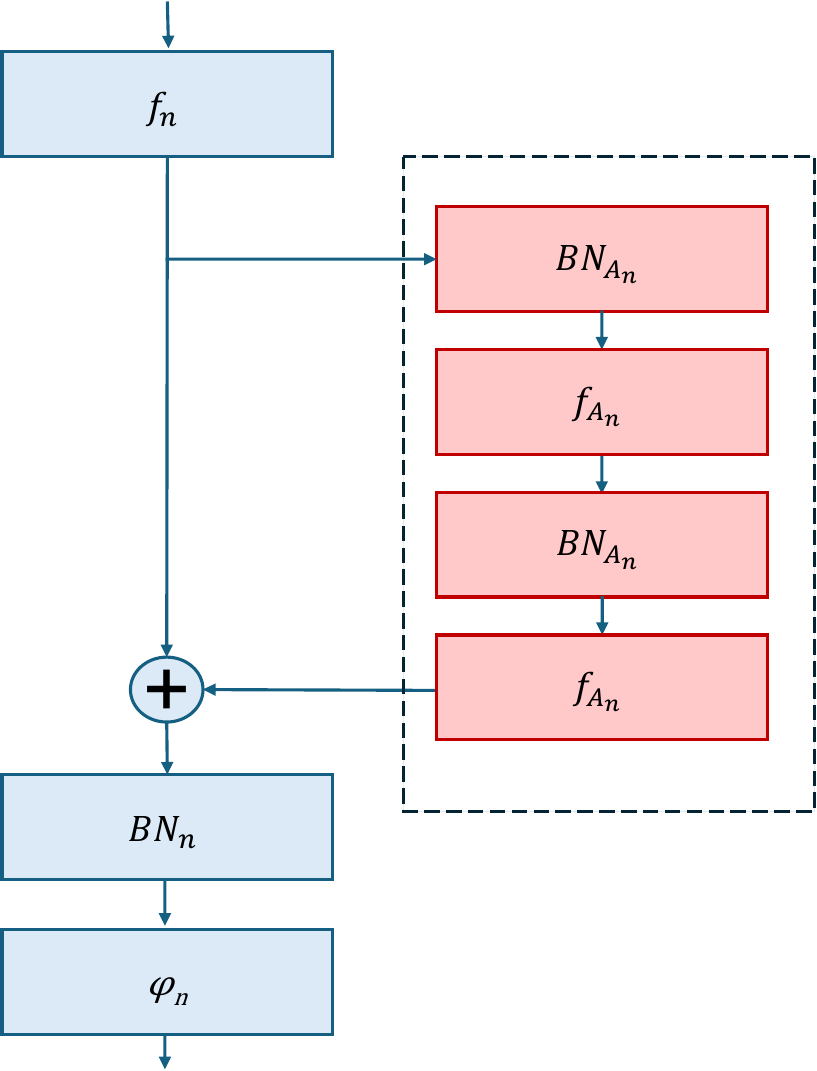}
     \caption{BN + Conv1x1 + BN + Conv1x1}
\end{subfigure}
\caption{Adapter designs schematics.}
\label{suppl_fig:adapters_schemes}
\end{figure*}

\newpage

\subsection{Gridsearch on baseline domain}
\label{suppl_sec:hyperparameters}
The outline of the baseline process was to study the synthetic dataset (\textit{SMo}), exploring different network parameters, such as:
\begin{itemize}[noitemsep]
    \item backbones
    \item optimizers
    \item loss functions
    \item data augmentations
\end{itemize}
All of these possible combinations were studied with the same training strategy: starting from a learning rate of $10^{-3}$, this will be kept the same until there are no improvements on the validation set for 10 epochs. If this happens, the learning rate will be cut by a factor of 10, for a maximum of 3 times. \\
Starting with the lightest ResNet backbone (i.e. Resnet-18), we explored different data augmentations, performing many combinations on our dataset; the studied augmentations are \textit{ColorJitter}, \textit{GaussianBlur}, \textit{HorizontalFlip}, \textit{RandomCrop} and \textit{Rotate}. Finally, the images were scaled using Gaussian normalization and transformed into grayscale images, proving that the three colour channels would have been uselessly redundant. As visible in the results Tables~\ref{suppl_tab:data_agumentations}, we noticed that the most significant one is the \textit{RandomCrop} transformation, bringing major improvements to the performances. \\
As the next step, we compared the \textit{Adam} optimizer with the \textit{SGDm}. Additionally, we studied different loss functions to find the best task-wise, such as the Balanced Categorical Cross-Entropy \textit{BCCE}, the Dice loss and the Jaccard loss, together with the matching metrics (Balanced Accuracy and mean Intersection-Over-Union scores). Notice that the balancing of the loss functions and the correspondent metrics is mandatory, due to the different class frequencies. The results are presented in  Tables~\ref{suppl_tab:optimizers_loss}, showing that the best choices are the Adam optimizer together with the BalancedCCE loss function.

\vspace{4cm}

\begin{table*}[ht]
\centering
\begin{tabular}{cc|c|c|c|c}
\multirow{ 2}{*}{Loss Function} & \multirow{ 2}{*}{Optimizer} & \multicolumn{2}{c|}{Balanced Accuracy} & \multicolumn{2}{c}{IoU}\\
 & & validation & test & validation & test \\
\hline
Jaccard & Adam & 93.86 & 91.37 & 90.11 & 87.34\\
Dice & Adam & 94.05 & 91.73 & 90.03 & 87.64\\
\hline
\textbf{BalancedCCE} & \textbf{Adam} & \textbf{96.49} & \textbf{94.98} & \textbf{81.27} & \textbf{82.32}\\
BalancedCCE & SGD & 96.00 & 94.51 & 82.31   & 80.48\\
\hline
\end{tabular}
\caption{Hyperparameter validation of optimizer and loss function using Resnet-18 encoder and random crop augmentation on \textit{SMo} dataset}
\label{suppl_tab:optimizers_loss}
\end{table*}

\newpage

\begin{table*}[ht]
\centering
\begin{tabular}{ccccc|c|c|c|c}
\multirow{ 2}{*}{RandomCrop} & \multirow{ 2}{*}{Rotate} & \multirow{ 2}{*}{HorizontalFlip} & \multirow{ 2}{*}{GaussianBlur} & \multirow{ 2}{*}{ColorJitter} & \multicolumn{2}{c|}{Balanced Accuracy} & \multicolumn{2}{c}{IoU}\\
 & & & & & validation & test & validation & test \\
\hline
No & No & No & No & No & 86.95 & 87.52 & 76.34 & 75.16 \\
No & Yes & No & No & No & 80.84 & 84.30 & 60.78 & 65.26\\
\textbf{Yes} & \textbf{No} & \textbf{No} & \textbf{No} & \textbf{No} & \textbf{96.49} & \textbf{94.98} & \textbf{83.76} & \textbf{82.32}\\
Yes & Yes & No & No & No & 95.90 & 94.45 & 80.94 & 79.20\\
No & No & Yes & No & No & 87.47 & 87.55 & 76.01 & 75.23\\
No & Yes & Yes & No & No & 77.00 & 81.83 & 55.78 & 61.21\\
Yes & No & Yes & No & No & 96.35 & 94.83 & 83.03 & 81.50\\
Yes & Yes & Yes & No & No & 95.76 & 94.31 & 80.63 & 78.80\\
No & No & No & Yes & No & 86.79 & 89.05 & 71.86 & 71.48\\
No & Yes & No & Yes & No & 79.83 & 84.53 & 58.12 & 63.24\\
Yes & No & No & Yes & No & 96.04 & 94.60 & 81.86 & 79.93\\
Yes & Yes & No & Yes & No & 95.54 & 94.13 & 79.82 & 78.11\\
No & No & Yes & Yes & No & 86.01 & 88.47 & 69.05 & 69.84\\
No & Yes & Yes & Yes & No & 79.48 & 84.44 & 57.94 & 63.14\\
Yes & No & Yes & Yes & No & 96.22 & 94.86 & 82.09 & 80.23\\
Yes & Yes & Yes & Yes & No & 95.43 & 94.02 & 79.75 & 77.60\\
No & No & No & No & Yes & 83.77 & 86.30 & 63.53 & 64.88 \\
No & Yes & No & No & Yes & 81.96 & 84.81 & 60.75 & 62.21\\
Yes & No & No & No & Yes & 95.97 & 94.47 & 81.86 & 80.16\\
Yes & Yes & No & No & Yes & 95.99 & 94.61 & 81.38 & 79.43\\
No & No & Yes & No & Yes & 88.02 & 88.58 & 68.02 & 66.92\\
No & Yes & Yes & No & Yes & 78.81 & 83.54 & 55.92 & 60.70\\
Yes & No & Yes & No & Yes & 96.03 & 94.62 & 81.55 & 79.56\\
Yes & Yes & Yes & No & Yes & 95.61 & 94.27 & 80.48 & 78.53\\
No & No & No & Yes & Yes & 88.58 & 90.32 & 72.84 & 72.43\\
No & Yes & No & Yes & Yes & 76.21 & 82.00 & 53.70 & 59.42\\
Yes & No & No & Yes & Yes & 95.99 & 94.67 & 81.62 & 80.13\\
Yes & Yes & No & Yes & Yes & 95.58 & 94.30 & 80.33 & 78.48\\
No & No & Yes & Yes & Yes & 84.10 & 87.10 & 64.53 & 66.70\\
No & Yes & Yes & Yes & Yes & 75.38 & 81.65 & 52.92 & 59.04\\
Yes & No & Yes & Yes & Yes & 95.71 & 94.37 & 80.54 & 78.58\\
Yes & Yes & Yes & Yes & Yes & 95.87 & 94.45 & 81.27 & 78.96\\
\hline
\end{tabular}
\caption{Grid search of augmentation technique using Resnet-18 encoder, Adam optimizer, and balanced cross-entropy loss function on \textit{SMo} dataset.}
\label{suppl_tab:data_agumentations}
\end{table*}

\newpage

\subsection{Preliminary Study on Traditional Approaches}
\label{sec:classic_alg_results}

\begin{figure*}[ht]
    \centering
    \includegraphics[width = \textwidth]{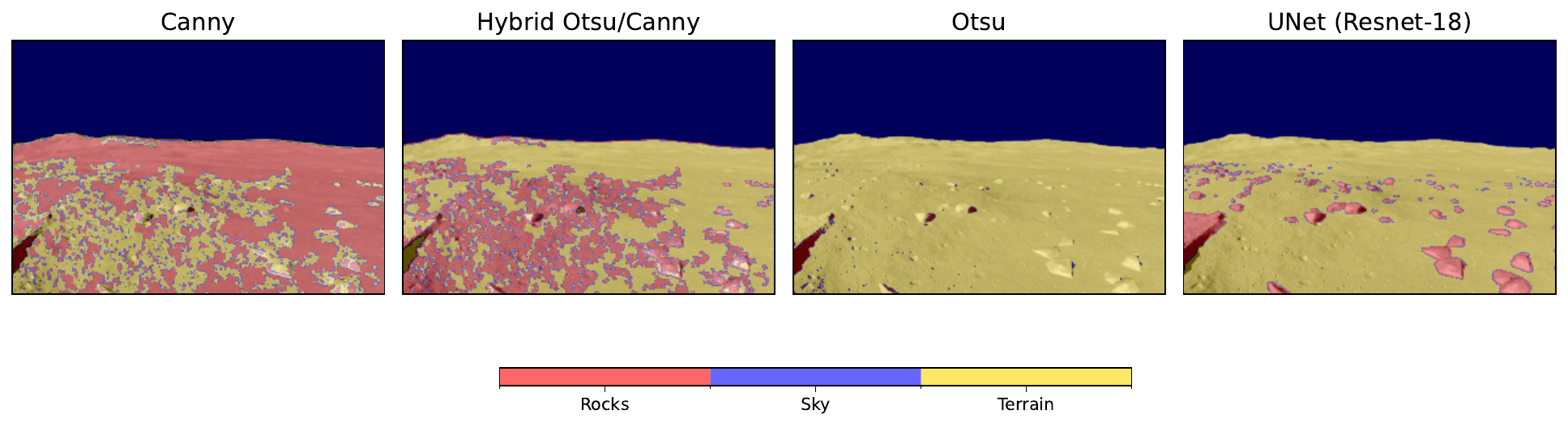}
    \caption{Comparison of traditional methods (\textit{Canny}, \textit{Otsu}, and \textit{Hybrid}) with DL results on the \textit{SMo} dataset.}
    \label{fig:classic_alg_comparison}
\end{figure*}

While this work focuses on Deep Learning (DL) techniques and their optimization, we also evaluated the baseline performances of these traditional algorithms on the same dataset to provide a fair comparison. As previously outlined by Thompson and Casta\~no~\cite{thompson2007performance}, traditional methods rely mainly on edge detection and thresholding for image segmentation. However, additional operations, such as contour classification, are required to assign detected regions to specific classes (e.g. distinguishing sky from terrain or rocks using criteria like contour size or threshold values). The primary advantage of these algorithms is their applicability to any dataset without requiring any training, eliminating its computational cost and the need for large datasets, which are significant for ML/DL approaches. Considering also the purpose of this work, this inherent adaptability makes them robust to changes in datasets, with performance depending solely on data quality and intrinsic characteristics and providing quick and automatic flexibility.\\
Despite these strengths, traditional methods generally underperform compared to DL models. In this work, for completeness, we implemented two well-known different algorithms on our three-classes datasets: the Canny edge detector~\cite{canny_alg} and the Otsu thresholding~\cite{otsu_alg}, plus we built a hybrid version of them to exploit their strengths.\\
Both of them are the base of the traditional computer vision approach, given that the former evaluates the gradient of the image, searching for steep changes, while the latter provides autonomous thresholding based on the image variance.
In the hybrid approach, \textit{Otsu} masks the background (sky and terrain), and \textit{Canny} refines rock detection. Note that these methods require grayscale images for processing. Results on the \textit{SMo} dataset are summarized in Table~\ref{tab:classic_alg}, with visual examples in Fig.~\ref{fig:classic_alg_comparison}. For consistency, the same train-validation-test split was used for DL methods, and we compared quantitative and qualitative results with a U-Net trained on the same datasets, using ResNet-18 as a backbone.

\begin{table*}[ht]
\centering
\caption{Performance of traditional algorithms (\textit{Canny}, \textit{Otsu}, and \textit{Hybrid Otsu/Canny}) compared with a U-Net with ResNet-18 as a backbone, on the \textit{SMo} and \textit{Marsdataset} datasets.}
\label{tab:classic_alg}
\begin{tabular}{c cc cc}
\toprule
\multirow{2}{*}{\textbf{Algorithm}} & \multicolumn{2}{c}{\textbf{Balanced Accuracy [\%]}} & \multicolumn{2}{c}{\bf IoU [\%]}\\
 & \textit{SMo} & \textit{Marsdataset}  & \textit{SMo} & \textit{Marsdataset} \\
\midrule
\textit{Otsu}   & 66.73 & 48.50 & 59.10 & 27.08 \\
\textit{Canny}  & 38.74 & 37.55 & 28.74 & 15.03 \\
\textit{Hybrid Otsu/Canny} & 67.59 & 45.98 & 45.74 & 20.01 \\
\textit{U-Net (ResNet-18)} & 94.98 & 86.96 & 82.32 & 50.72 \\
\bottomrule
\end{tabular}
\end{table*}

As observed in Table~\ref{tab:classic_alg} and Fig.~\ref{fig:classic_alg_comparison}, traditional methods struggle with rock segmentation. While \textit{Otsu} effectively identifies terrain and sky, it often detects only rock shadows. On the other hand, \textit{Canny} identifies better the rocks' edges but fails to capture the shapes accurately. 
Results on the Martian dataset further highlight the limitations of traditional methods. In this case, the different physical environments play a main role: the Moon's high contrast (dark sky and grayscale landscape) supports better the segmentation of these techniques. On the other hand, Mars' atmospheric effects, such as dust storms and sunlight scattering, reduce the contrast between sky and terrain, making more difficult the edges and background distinctions and leading to significantly worse performance. On these setups, a basic U-Net solution largely outperforms traditional methods, improving all the metrics by at least 25\%, suggesting that the great performance of a DL-based solution justifies a larger computation overhead.

\newpage

\subsection{Visualization of Adapter Ranking (Vgg19-BN)}



\begin{figure*}[ht]
    \centering
    \includegraphics[scale=0.4]{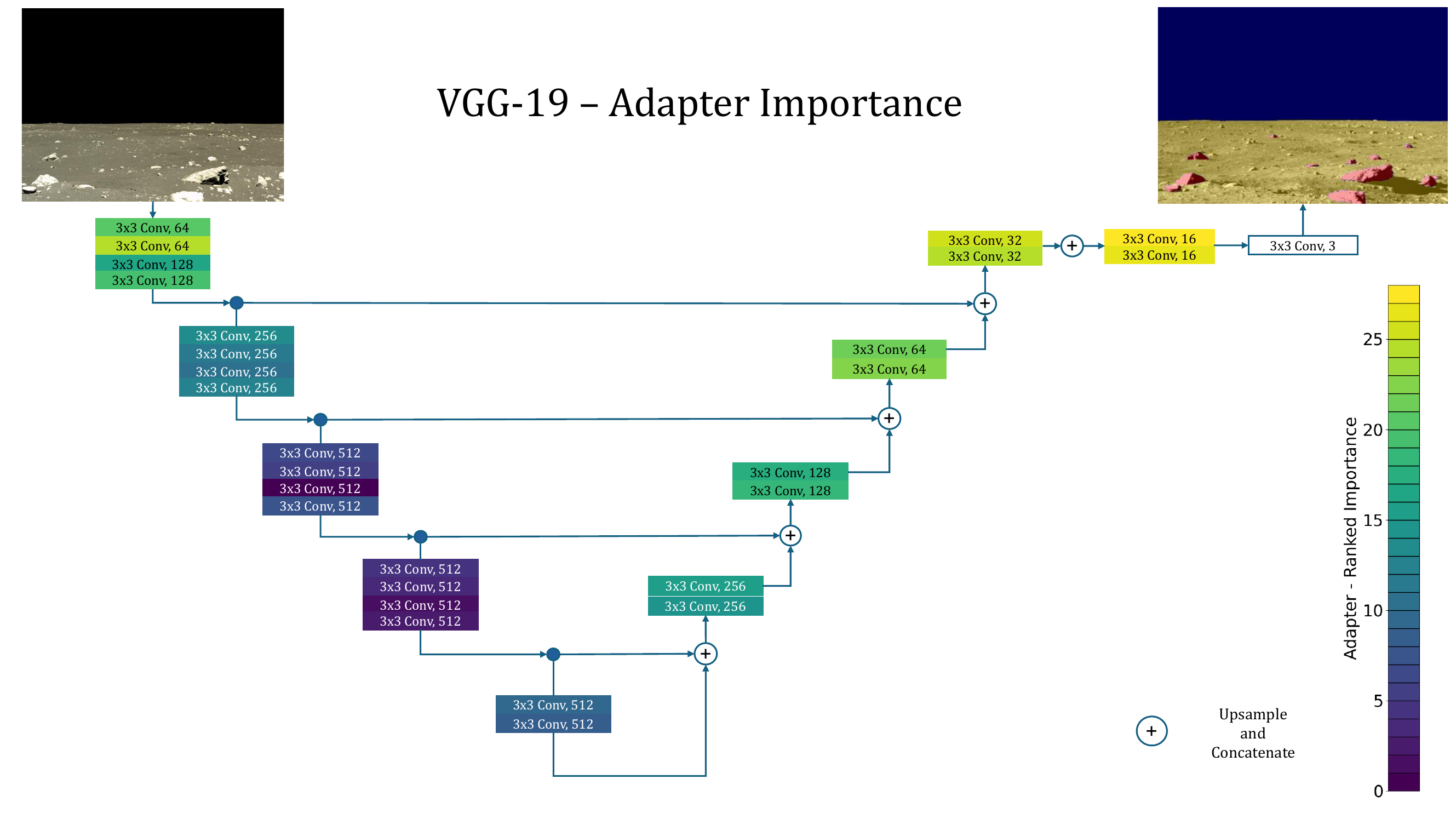}
    \caption{Visualization of Vgg-19 adapter ranking: the least important modules are in the encoder layer just before the bottomneck. The most important ones are the first encoder layers and the last decoder layers.}
    \label{suppl_fig:vgg19_ranking}
\end{figure*}


\vspace{2.5cm}
\subsection{Noise robustness}
\label{suppl_sec:noise}

\begin{figure}[ht]
\centering
\begin{subfigure}[b]{0.32\textwidth}
     \centering
     \includegraphics[scale=0.38]{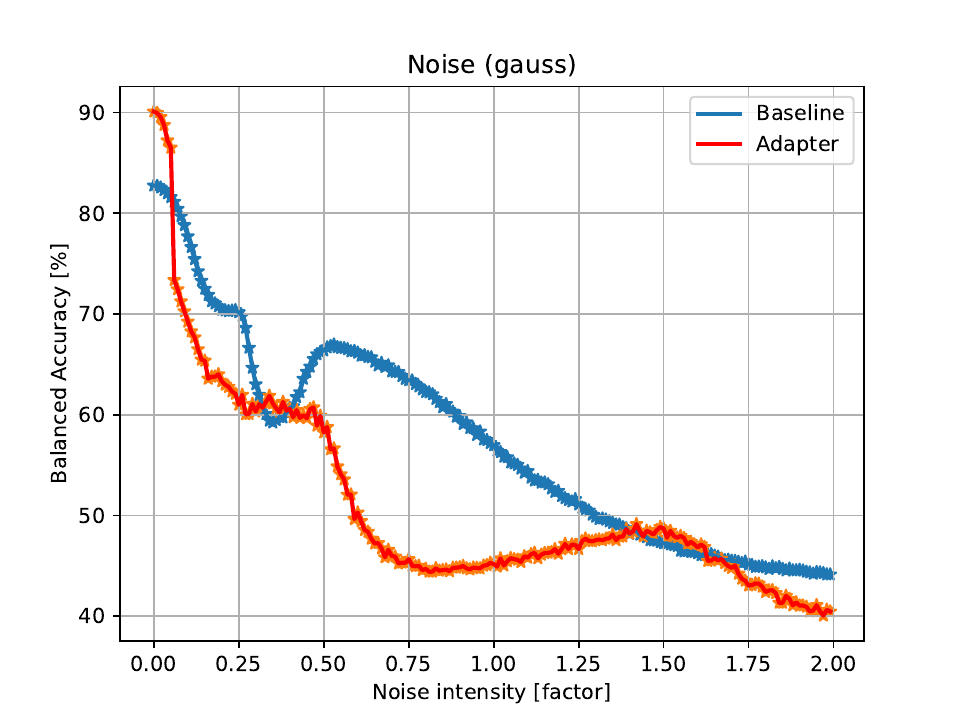}
     \caption{Gaussian}
     \label{subfig:gauss}
\end{subfigure}
\begin{subfigure}[b]{0.32\textwidth}
     \centering
     \includegraphics[scale=0.38]{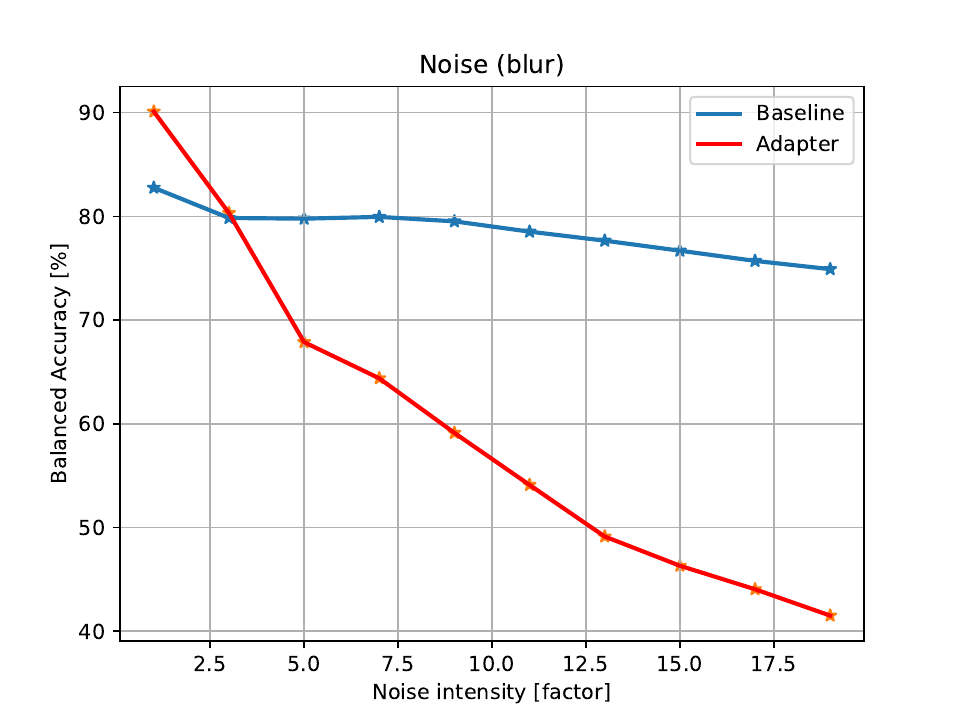}
     \caption{Blurring}
      \label{subfig:blurr}
\end{subfigure}
\begin{subfigure}[b]{0.32\textwidth}
     \centering
     \includegraphics[scale=0.38]{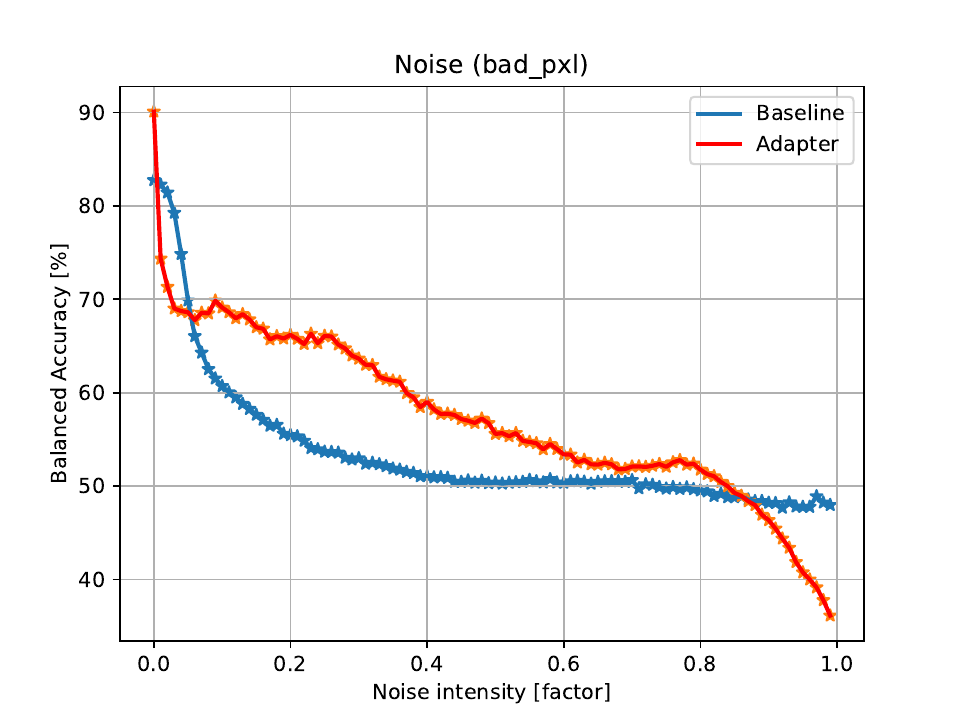}
     \caption{Pixel malfunctioning}
      \label{subfig:bad_pxl}
\end{subfigure}
\caption{Noise robustness for Vgg19-BN backbone on \textit{RMo} for different type of noise: "gauss" = white noise on image, "blur" = out-of-focus optics, "bad\_pxl" = pixel malfunctioning.}
\label{fig:noise}
\end{figure}

Autonomous rovers operating in unstructured environments are subject to various sources of image degradation that can impact perception and decision-making. We evaluate the robustness of our models to three different kinds of realistic noise, even comparing whether the introduction of the adapter amplifies them or not. The three realistic noise conditions are a) Gaussian noise (\ref{subfig:gauss}), simulating sensor white disturbance, b) blurring (\ref{subfig:blurr}), representing out-of-focus optics, and c) pixel malfunctions (\ref{subfig:bad_pxl}), mimicking faulty sensor elements. The following results are evaluated on the \textit{RealMoon} dataset and evaluating the Vgg19-BN backbone architecture, as a preliminary example.\\
The three plots show the comparison between the performance of a baseline model and an adapter-equipped one, as a function of the intensity of the related noise. The latter differs for each noise: for Gaussian intensity, it means the white noise apmòitude ($\in [0,\inf]$); regarding the blurring, it represents out-focus kernel size (odd numbers), while about pixel malfunctioning, it is a percentage of the "broken" (i.e. turned off) pixels on the total ($\in [0,1]$). Across all noise types, the balanced accuracy decreases as the noise intensity increases, confirming the expected degradation in performance. The initial drop is quite quick (for both of the models), meaning that they are not quite robust to these kinds of noises. In the case of Gaussian noise, the adapter-equipped model drops significantly against a smoother behaviour of the baseline. This one is relatively robust to the blurring noise, instead of the adapter one, which drops its performance consistently with the noise intensity. Regarding the pixel malfunctioning, the adapter model behaves better than the baseline: despite the common initial drop, the performance is higher than the baseline pretty much up to $80\%$ of the corrupted pixels.


\newpage

\subsection{Layer by layer finetuning}
\label{suppl_sec:layerbylayer_ablation}

\begin{figure}[t]
    \centering
    \includegraphics[scale=0.65]{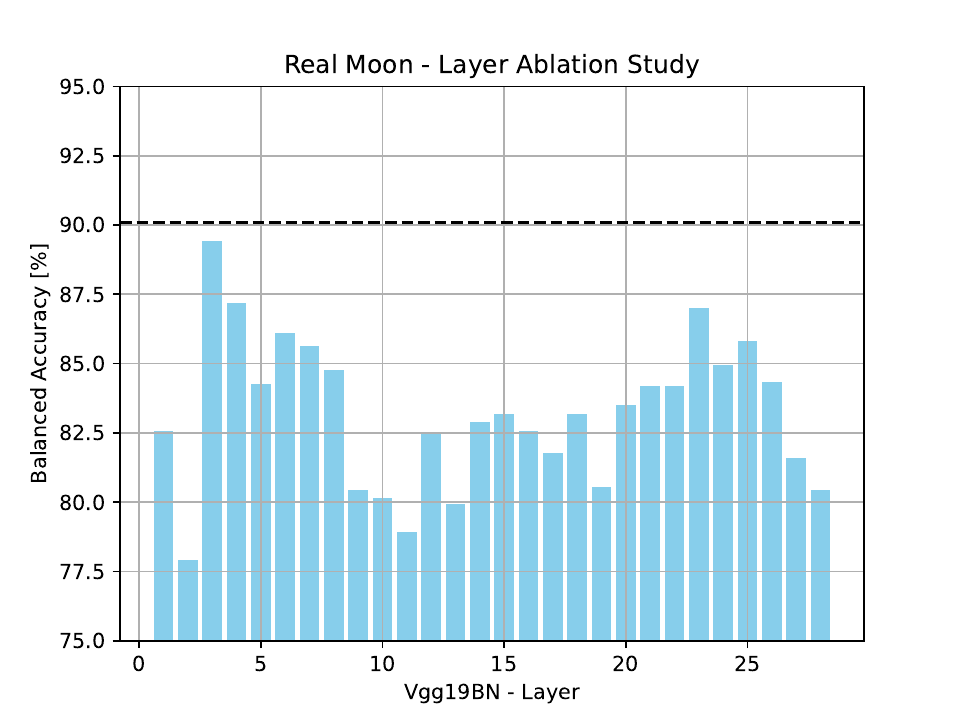}
    \caption{Balanced Accuracy as a function of the fine-tuned layer for the Vgg19-BN backbone model on \textit{RMo}.}
    \label{fig:layerbylayer}
\end{figure}

A way to explore the best place to make the update, the adapter insertion, for example, could be to perform a layer-by-layer fine-tuning, measuring their performance. To investigate the impact of this layer-wise fine-tuning, we conducted a systematic ablation study, using the Vgg19-BN backbone on the RealMoon dataset as a case study. Beginning with a fully frozen network, we progressively fine-tuned each layer, moving sequentially in the architecture, while evaluating balanced accuracy at each step.\\
The resulting plot, see Fig.~\ref{fig:layerbylayer} illustrates the effect of fine-tuning specific layers on model performance. Our findings reveal that fine-tuning the early convolutional layers (i.e. the encoder ones) yields the main improvements, together with the last ones (i.e. the decoder ones), although to a lesser extent. The difference in performance between these and the central ones is quite significant ($>5\%$) and resembles the results presented in the ``Adapter ranking" (see Sec.~\ref{sec:adapter_ranks_esec:xperiments}). This confirms the adapter ranking results regarding the update positioning, while also showing the importance of the lightest layers of the UNet architecture for feature extraction.

\newpage

\subsection{Embedded devices}
\label{suppl_sec:devices}
The Raspberry~Pi~4 mounts a 1.8 GHz quad-core CPU and 8 GBytes of RAM.
The Jetson Orin Nano has a 625 MHz 1024-core GPU and 8 GBbytes of RAM.
Regarding their performances, as expected, the inference times of the JetsonNano are much lower than the Raspberry, due to the different hardware. The former ranges between $\approx0.6$ and $\approx1.4$ seconds, while the latter shows a significant increase.
In particular, execution times of ResNet-18 U-Nets vary between $\approx 6.9$  seconds of a base architecture to $\approx 10.1$  seconds with adapters included. Times are also higher for the Vgg19-BN architecture, which requires much more computational power, with execution times ranging from $\approx20.5$ to $\approx26.7$ seconds. 
A similar pattern can be observed in FLOPs measurements, with a mean increase of $\approx 10\%$ for adapter-injected U-Nets. 
Besides, equipped adapter models increase (reaching even six times) the memory consumption, especially affecting the Vgg19 ones. This limits the image size that can be processed as input, which cannot overcome the $1024\times1024$ pixels size on these devices. That's the reason why in the AI4Mars dataset we had to rescale the images by a factor of 2. 
Finally, the advantage of using adapters becomes clear considering the storage memory measurements, specifically their compressed counterparts. The storage required is significantly smaller than base architectures, weighing between $11\%$ for ResNet-18 and $20\%$ for Vgg19-BN of the entire base model.

\newpage

\begin{table*}
\centering
\caption{Performance on Raspberry Pi 4.}
\begin{adjustbox}{max width=\textwidth}
\begin{tabular}{@{}cccccccccc@{}}
\toprule
\multirow{3}{1.2cm}{\centering \bf Device} & \multirow{2}{*}{\bf Dataset} & \multirow{3}{*}{\bf Encoder} & \multirow{3}{*}{\bf Method} & \multirow{2}{1.8cm}{\centering \bf Balanced \\ Accuracy} & \multirow{2}{*}{\bf Time } & \multirow{2}{1.4cm}{\centering \bf Memory \\ usage }  & \multirow{2}{*}{\bf FLOPs } & \multirow{2}{1.8cm}{\centering\bf Storage\\ Memory } & \multirow{2}{1.8cm}{\centering\bf Compressed \\(.tar.gz) } \\
 & & & & & & & & & \\
 & \bf (size) & & & \bf (\%) & \bf [sec] & [Mb] & \bf [G] & [MB] & [MB]\\
 
\midrule
\multirow{30}{*}{\rotatebox[origin=c]{90}{\Large Raspberry Pi 4}} &  \multirow{10}{1.4cm}{\centering RMo \\ (480,704)} & \multirow{5}{*}{ResNet-18} & baseline & 82.14 & \multirow{3}{*}{8.487} & \multirow{3}{*}{718.96} & \multirow{3}{*}{27.5} & \multirow{3}{*}{54.75} & \multirow{3}{*}{50.78} \\     
 &   &  &  scratch & 85.41 &   &   &   &  & \\
 &   &  &   finetuning & 92.70 &  &  &   &   &  \\
\cmidrule{4-10}
 &   &  &  adapter & 89.68 & 10.186 & 1229.76 & 30.17 & +6.12 & +5.64 \\
   &   &  &  fused adapter & 90.61 & 8.646	& 641.63 & 27.42 & 54.67 & 50.80 \\
 \cmidrule{4-10}
 &   & \multirow{5}{*}{Vgg19-BN} & baseline & 70.08 & \multirow{3}{*}{25.775} & \multirow{3}{*}{1506.23} & \multirow{3}{*}{155.54} & \multirow{3}{*}{111.01} & \multirow{3}{*}{55.00} \\
 &   &  &  scratch & 82.76 &   &   &   &  & \\
 &   &  & finetuning & 93.65 &  &  & &   & \\
\cmidrule{4-10}
 &   &  & adapter & 90.10 & 33.259 & 2804.3 & 174.81 & +11.99 & +11.04\\
  &   &  &  fused adapter & 87.81 & 25.86 & 1096.71 & 155.3 & 110.88 & 103.03 \\
\cmidrule{2-10}
\cmidrule{2-10}

 & \multirow{10}{1.4cm}{\centering AI4Mars \\ (512,512)} & \multirow{5}{*}{ResNet-18} & baseline & 34.05 & \multirow{3}{*}{6.991} & \multirow{3}{*}{563.11} & \multirow{3}{*}{21.33} & \multirow{3}{*}{54.75} & \multirow{3}{*}{50.78}\\
  &   &  &  scratch & 87.89 &   &   &   &  & \\
 &   &  &    finetuning & 90.53 &  &  &   &   & \\
\cmidrule{4-10}
 &   &  &  adapter & 89.99 & 8.348 & 959.81 & 23.4 & +6.12 & +5.64\\
   &   &  &  fused adapter & 80.30 & 7.045	& 492.31 & 21.27 & 54.67 & 50.80 \\
\cmidrule{3-10}
 &   & \multirow{5}{*}{Vgg19-BN} & baseline & 35.21 & \multirow{3}{*}{20.517} & \multirow{3}{*}{1163.95} & \multirow{3}{*}{120.66} & \multirow{3}{*}{111.01} & \multirow{3}{*}{55.00}\\
 &   &  &  scratch & 86.71 &   &   &   &  & \\
 &   &  & finetuning & 90.34 &  &  & &   & \\
\cmidrule{4-10}
 &   &  & adapter & 90.69 & 26.672 & 2159.09 & 135.61  & +11.99 & +11.04\\
   &   &  &  fused adapter & 86.18 & 20.461 & 883.52 & 120.47 & 110.88 & 103.03 \\
\cmidrule{2-10}
\cmidrule{2-10}

 & \multirow{10}{1.9cm}{\centering MarsDataset \\ (512,512)} & \multirow{5}{*}{ResNet-18} & baseline & 57.61 & \multirow{3}{*}{6.964} & \multirow{3}{*}{564.58} & \multirow{3}{*}{21.33} & \multirow{3}{*}{54.75} & \multirow{3}{*}{50.78}\\
  &   &  &  scratch & 86.96 &   &   &   &  & \\
 &   &  & finetuning & 93.46 &   &   &   &  &\\
\cmidrule{4-10}
 &   &  & adapter & 91.53 & 8.294 & 965.55 & 23.4 & +6.12 & +5.64\\
  &   &  &  fused adapter & 89.48 & 7.051 & 472.83 & 21.27 & 54.67 & 50.80 \\
  \cmidrule{3-10}
 &   & \multirow{5}{*}{Vgg19-BN} & baseline & 53.54 & \multirow{3}{*}{20.665} & \multirow{3}{*}{1184.78} & \multirow{3}{*}{120.66} & \multirow{3}{*}{111.01} & \multirow{3}{*}{55.00} \\
 &   &  &  scratch & 83.92 &   &   &   &  & \\
 &   &  & finetuning & 91.64 &  &  & &  &\\
\cmidrule{4-10}
 &   &  & adapter & 87.96 & 26.489 & 2167.91 & 135.61 & +11.99 & +11.04\\
  &   &  &  fused adapter & 87.81 & 20.611 & 849.54 & 120.47 & 110.88 & 103.03 \\
\midrule
\midrule

\end{tabular}
\end{adjustbox}
\label{suppl_tab:rasppi}
\end{table*}

\begin{table*}
\centering
\caption{Performance on Jetson Nano Orin.}
\begin{adjustbox}{max width=\textwidth}
\begin{tabular}{@{}cccccccccc@{}}
\toprule
\multirow{3}{1.2cm}{\centering \bf Device} & \multirow{2}{*}{\bf Dataset} & \multirow{3}{*}{\bf Encoder} & \multirow{3}{*}{\bf Method} & \multirow{2}{1.8cm}{\centering \bf Balanced \\ Accuracy} & \multirow{2}{*}{\bf Time } & \multirow{2}{1.4cm}{\centering \bf Memory \\ usage }  & \multirow{2}{*}{\bf FLOPs } & \multirow{2}{1.8cm}{\centering\bf Storage\\ Memory } & \multirow{2}{1.8cm}{\centering\bf Compressed \\(.tar.gz) } \\
 & & & & & & & & & \\
 & \bf (size) & & & \bf (\%) & \bf [sec] & [Mb] & \bf [G] & [MB] & [MB]\\

\multirow{30}{*}{\rotatebox[origin=c]{90}{\Large Jetson Orin Nano}}  &  \multirow{10}{1.4cm}{\centering RMo \\ (480,704)} & \multirow{5}{*}{ResNet-18} & baseline & 82.14 & \multirow{3}{*}{0.639} & \multirow{3}{*}{113.57} & \multirow{3}{*}{27.5} & \multirow{3}{*}{54.75} & \multirow{3}{*}{50.78}\\
 &   &  &  scratch & 85.41 &   &   &   &  & \\
 &   &  &  finetuning & 92.70 &   &   &   &  & \\
\cmidrule{4-10}
 &   &  &  adapter & 89.68 & 0.785 & 331.97 & 30.17 & +6.12  &  +5.64\\
   &   &  &  fused adapter & 90.61 & 0.701 & 108.1 & 27.42 & 54.67 & 50.80 \\
\cmidrule{3-10}
 &   & \multirow{5}{*}{Vgg19-BN} & baseline & 70.08 & \multirow{3}{*}{0.951} & \multirow{3}{*}{636.83} & \multirow{3}{*}{155.54} & \multirow{3}{*}{111.01} & \multirow{3}{*}{55.00}\\
 &   &  &  scratch & 82.76 &   &   &   &  & \\
 &   &  & finetuning & 93.65 &   &   &   &  & \\
\cmidrule{4-10}
 &   &  & adapter & 90.10 & 1.452 & 2241.03 & 174.81 & +11.99  & +11.04\\
&   &  &  fused adapter & 87.81 & 0.967 & 645.18 & 155.3 & 110.88 & 103.03 \\
\cmidrule{2-10}
\cmidrule{2-10}

 &\multirow{10}{1.4cm}{\centering AI4Mars \\(512,512)} & \multirow{5}{*}{ResNet-18} & baseline & 34.05 & \multirow{3}{*}{0.603} & \multirow{3}{*}{118.81} & \multirow{3}{*}{21.33} & \multirow{3}{*}{54.75} & \multirow{3}{*}{50.78}\\
  &   &  &  scratch & 87.89 &   &   &   &  & \\
 &   &  & finetuning & 90.53 &    &   &   &  & \\
\cmidrule{4-10}
 &   &  &  adapter & 89.99 & 0.802 & 318.27 & 23.4 & +6.12 & +5.64\\
&   &  &  fused adapter & 80.30 & 0.601	& 115.48 & 21.27 & 54.67 & 50.80 \\
\cmidrule{3-10}
 &   & \multirow{5}{*}{Vgg19-BN} & baseline & 35.21 & \multirow{3}{*}{0.781} & \multirow{3}{*}{474.23} & \multirow{3}{*}{120.66} & \multirow{3}{*}{111.01} & \multirow{3}{*}{55.00}\\
 &   &  &  scratch & 86.71 &   &   &   &  & \\
 &   &  & finetuning & 90.34 &   &   &   &  &  \\
\cmidrule{4-10}
 &   &  &  adapter & 90.69 &	1.298 & 1762.71 & 135.61 & +11.99  & +11.04\\
   &   &  &  fused adapter & 86.18 & 0.775 & 552.12 & 120.47 & 110.88 & 103.03 \\
\cmidrule{2-10}
\cmidrule{2-10}

 &\multirow{10}{1.9cm}{\centering MarsDataset \\(512, 512)} & \multirow{5}{*}{ResNet-18} & baseline & 57.61 & \multirow{3}{*}{0.601} & \multirow{3}{*}{100.36} & \multirow{3}{*}{21.33} & \multirow{3}{*}{54.75} & \multirow{3}{*}{50.78} \\
  &   &  &  scratch & 86.96 &   &   &   &  & \\
 &   &  & finetuning & 93.46 &   &   &   &  & \\
\cmidrule{4-10}
 &   &  &  adapter & 92.78 & 0.77 & 115.28 & 23.4 & +6.12 & +5.64\\
  &   &  &  fused adapter & 89.48 & 0.634 & 101.40 & 21.27 & 54.67 & 50.80 \\
\cmidrule{3-10}
 &   & \multirow{5}{*}{Vgg19-BN} & baseline & 53.54 & \multirow{3}{*}{0.813} & \multirow{3}{*}{309.51} & \multirow{3}{*}{120.66} & \multirow{3}{*}{111.01} & \multirow{3}{*}{55.00}\\
 &   &  &  scratch & 83.92 &   &   &   &  & \\
 &   &  & finetuning & 91.64 &   &   &   &  &  \\
\cmidrule{4-10}
 &   &  &  adapter & 87.96 & 1.335 &	1768.57 & 135.61 & +11.99 & +11.04\\
  &   &  &  fused adapter & 87.81 & 0.826 & 258.37 & 120.47 & 110.88 & 103.03 \\

\midrule

\end{tabular}
\end{adjustbox}
\label{suppl_tab:jnano}
\end{table*}

%% file: suppl_data.tex
\subsection{Data}
\label{suppl_sec:data}

In this section, we present in detail the studied datasets together with their issues and the required pre-processing.\\
The studied datasets are three, provided by NASA's open-source database. The first one, called \textit{Synthetic Moon} dataset, contains artificial renders of the lunar landscape, together with some real photos of the Moon's environment (\textit{Real Moon}). The other two regarding Mars are named \textit{AI4Mars} and \textit{MarsData-V2} and consist of real photographs of the Mars landscape. 

\subsubsection{Synthetic Moon Dataset  (\textit{SMo})}
The Synthetic Moon dataset \textit{SMo} (see \cite{smob} for details) currently consists of $9.766$ realistic renders of rocky lunar landscapes and their segmented equivalents (the 3 classes are the \textit{sky}, \textit{smaller rocks} and \textit{larger rocks}). The resolution of the images is $480\times720\times3$; the $\times3$ reflects three channels of RGB colours. This dataset was created by Romain Pessia and Genya Ishigami of the Space Robotics Group, Keio University, Japan, using NASA's \textit{LRO LOLA Elevation Model} as a source of large-scale terrain data. The dataset provides two kinds of segmented maps: a "ground" map and a "clean" one. The latter is a rougher version of the first one, cutting a lot of small segmented pixels region. Thus, for the analysis, the "ground" mask will be used because there are more mask labels for rocks. 

\subsubsection{Real Moon (\textit{RMo})}
This Moon dataset also contains 36 real photographs of lunar landscapes and correspondent ground truth equivalent (same classes). We will refer to these images as the Real Moon dataset (\textit{RMo}). The resolution of the images is $571\times777\times3$, except for one image (PCAM14) having $514\times700\times3$ resolution and five (TCAM1,TCAM2,TCAM3,TCAM4,TCAM5) with $1728\times2352\times3$ resolution. We consider two different Mars datasets for the domain adaptation task, due to some critical issues of the first one \textit{AI4Mars} discovered in progress: its labelling is unreliable and not coherent within itself. Because of this, we proceed to use a newer and more reliable one, the \textit{MarsData-V2}. It is smaller but with better labels about rocks and terrain.

\subsubsection{AI4Mars}
AI4Mars dataset consists of $\sim18100$ images from Curiosity, Opportunity, and Spirit rovers, collected through crowdsourcing (see \cite{AI4Mars} for more details). Each image was labelled by 10 people to ensure greater quality and agreement with the crowdsourced labels. The dataset has already been divided into train and test sets: the first one contains a cleaned and merged version of crowdsourced labels with a minimum agreement of 3 labellers and $2/3$ agreement for each pixel; furthermore distances further than 30 meters and the rover itself are masked. The test set contains a merged version of expert-gathered labels with $100\%$ agreement required on each pixel. Three different versions are provided specifying the minimum number of people agreeing on each pixel labelling (\textit{"min3"} is the sparsest as it requires all 3 labellers to agree and we are going to use it). Furthermore, as said before, the dataset provides also a rover mask and distance-over-30-meters mask for each image (binary masks). An important thing to underline is that here the labels are different: 5 classes, \textit{soil}, \textit{bedrock}, \textit{sand}, \textit{big rocks} and 255 \textit{NULL} class (no label). This difference from the Moon dataset will be explored in the following section. The resolution of the images in the Martian dataset is $1024\times1024$; the $\times3$ channels to match the ResNet requirements are artificially created and are all the same, while the classes are coded as different colour values (0 \textit{soil}, 1 \textit{bedrock}, 2 \textit{sand}, 3 \textit{big rocks} and 255 \textit{NULL}) on the image.


\subsubsection{MarsData-V2}
\textit{MarsData-V2} (see \cite{marsdata}) is a rock segmentation dataset of real Martian scenes extended from a previously published \textit{MarsData}. The raw RGB images were collected by a Mastcam camera of the Curiosity rover on Mars between August 2012 and November 2018, courtesy of NASA/JPL-Caltech (see NASA policy \url{https://www.jpl.nasa.gov/jpl-image-use-policy}). The dataset consists of 835 labelled images with a resolution of $512\times512$, divided into 504 images for training, 168 for validation, and 167 for testing. The labelling refers just to rocks and terrain.

\subsubsection{Merging labels across domains}
The first thing to do is a common annotation between the domains (i.e. datasets): the classes must be the same between each of them to be able to adapt the starting NN on different images. A reasonable application of this work (with the necessary modifications) is autonomous driving for rover missions, so, due to the already available classifications and this goal the choice is to have three different classes: \textit{rocks}, the major obstacle for a rover's route, \textit{terrain}, the drivable ground for a rover (apart from sand), and the \textit{sky}. \\
Regarding the moon dataset, two of the three chosen classes are already present in the dataset: rocks and sky. However, bigger rocks are distinguished from smaller ones, so these two classes are merged into one, due to our porpuses. The \textit{terrain} class is built from all of the other non-classified pixels. In case of superimposition between sky and rocks pixels (occurred sometimes), we decided to prefer the rocks over the sky, prioritizing the danger of rocks misclassification over a sky one (the sky is far away from the rover position and does not represent a danger to it). This procedure is performed both on \textit{SMo} and \textit{RMo}. An example is shown in figure \ref{suppl_fig:moon-merging}.
An analogue procedure is mandatory on the Mars dataset. The starting one was \textit{AI4Mars}: this contains 5 different classes and, in this case, the merging would be more complex. The \textit{soil}, \textit{bedrock}, and \textit{sand}, together with all \textit{NULL} pixels, constitute the new \textit{terrain} class, while the already existent \textit{big rocks} becomes the new \textit{rocks} one. Due to the fact that there's no clear labelling of the horizon and the presence of the \textit{over-30m} mask, the latter becomes the \textit{sky} class. However, this labelling is often unprecise and not coherent between the images. An example is shown in figure \ref{suppl_fig:mars-merging}.
As stated before, for this reason, another Martian dataset is chosen, \textit{MarsData-V2}. In this case, the existing labels are just two: \textit{rocks} and the rest. Since the absence of the sky classes, we decided to manually label the sky pixels, so in the end, the desired classes are reached: \textit{rocks} already exists, \textit{sky} is manually added and all of the remaining pixels become the new \textit{terrain} class. \\
Finally, due to the different number of images in the datasets, the division between training, validation, and test datasets is independent between the three domains. The leading criterium is the same: a bigger training dataset, while a similar size for the validation and testing ones.